\documentclass{article} 
\usepackage{iclr2027_conference,times}

\usepackage{amsmath,amsfonts,bm}

\def\eqref#1{equation~\ref{#1}}

\def\1{\bm{1}}

\DeclareMathAlphabet{\mathsfit}{\encodingdefault}{\sfdefault}{m}{sl}
\SetMathAlphabet{\mathsfit}{bold}{\encodingdefault}{\sfdefault}{bx}{n}

\newcommand{\softmax}{\mathrm{softmax}}

\usepackage[hidelinks]{hyperref}
\usepackage{url}

\usepackage{url}            
\usepackage{booktabs}       
\usepackage{amsfonts}       
\usepackage{nicefrac}       
\usepackage{microtype}      
\usepackage{xcolor}         
\usepackage{capt-of}
\usepackage[toc,page]{appendix}
\usepackage{titletoc}
\usepackage{amsmath}
\usepackage{multirow}
\usepackage{wrapfig} 
\usepackage{comment}
\usepackage{graphicx}
\usepackage{colortbl}
\usepackage{setspace}
\newcommand{\lp}[1]{\textcolor{red}{#1}}
\usepackage[normalem]{ulem} 
\usepackage{subcaption}

\usepackage{algorithm}
\usepackage{algpseudocode}

\newcommand{\std}[1]{\,\scriptsize{\textpm{}#1}}

\title{Masked Generative Motion Planning with Geometry-Guided Token Search}

\author{
\begin{tabular}{ccc}
Lipeng Zhuang\thanks{Equal contribution. 
\texttt{\{2658047z, 2649534r\}@student.gla.ac.uk}} &
Yingdong Ru\footnotemark[1] &
Shiyu Fan\thanks{Corresponding author. 
\texttt{shiyu.fan@glasgow.ac.uk}}
\thanks{Equal supervision.}\footnotemark[3] \\
Edmond S. L. Ho &
Gerardo Aragon Camarasa &
Paul Henderson\footnotemark[3] \\
\multicolumn{3}{c}{University of Glasgow, Glasgow, United Kingdom}
\end{tabular}
}

\iclrfinalcopy
\begin{document}

\maketitle

\fancyhead{}
\renewcommand{\headrulewidth}{0pt}

\begin{abstract}

Generative motion planners typically use learned trajectory priors for initial generation, while leaving test-time repair to local continuous refinement.
We introduce Masked Generative Motion Planning (MGMP), which extends the learned prior from efficient parallel generation to structural repair.
A masked generative transformer generates discrete trajectory candidates in parallel, and \textit{Geometry-Guided Token Search (GGTS)} uses scene geometry to target where to edit and which prior-supported alternatives to evaluate.
This turns refinement into an efficient search over discrete motion alternatives, enabling route-level restructuring beyond local trajectory deformation.
MGMP achieves 96\% success on \textit{Ring Maze} and 82\% repair success on \textit{Controlled Route Invalidation on Kuka}, exceeding the strongest external baselines by 23 and 25 percentage points, respectively.
It further generalizes to unseen layouts, additional obstacles, unseen geometries, single- and dual-arm planning, and real-world Baxter tasks.

\end{abstract}

\section{Introduction}
Motion planning in robotics aims to find a feasible trajectory between a start and goal configuration while satisfying geometric constraints~\citep{lavalle2006planning}. Trajectory-optimization methods directly refine trajectories against task-specific objectives~\citep{ratliff2009chomp,schulman2014motion}, but their performance is typically sensitive to initialization in cluttered and non-convex environments~\citep{pan2014predicting}. Generative planners mitigate this sensitivity by learning multimodal trajectory priors and producing diverse candidate motions that can subsequently be refined to satisfy collision constraints.
Generative motion planning therefore involves two coupled inference problems: \textit{efficient initial trajectory generation} and \textit{effective test-time geometric refinement}.

Existing planners make trade-offs on both fronts. For \emph{initial generation}, diffusion-based methods rely on iterative denoising in a high-dimensional continuous space~\citep{janner2022planning,carvalho2023motion}, leading to high inference cost; fewer denoising steps improve efficiency but can reduce sample quality.
For \emph{test-time refinement}, generated trajectories are typically refined through continuous guidance or trajectory optimization~\citep{carvalho2023motion,luopotential,yang2026safeflowmatcher}. Weak guidance may leave collisions unresolved, while stronger guidance can move trajectories away from the learned prior~\citep{yang2026safeflowmatcher}.
Such continuous refinement is effective when feasibility can be recovered through local adjustments; however, in harder cases such as near-enclosing obstacles, reaching a feasible alternative may require many iterations, remain sensitive to initialization and guidance strength, and still fail to escape unfavorable local basins.~\citep{pan2014predicting,luopotential}.

We therefore ask: \textit{How can a learned trajectory prior directly expose qualitatively different routes as candidates for structural repair, rather than expecting continuous refinement to reach them?}


We introduce Masked Generative Motion Planning (MGMP), which addresses this question by formulating motion planning as masked generation over discrete trajectory tokens, reusing the same token space for test-time structural repair.
A trajectory tokenizer learns compact motion tokens from feasible trajectories. For initial generation, a Masked Generative Transformer (MGT)~\citep{chang2022maskgit,chang2023muse} predicts token sequences in parallel, producing high-quality trajectory candidates in a single forward pass and reducing the need for costly resampling and post-hoc repair during online planning. Importantly, the same representation allows us to cast repair as a guided search over prior-supported motion alternatives, instead of continuously deforming the current trajectory. 

To make this search effective, we introduce \textbf{Geometry-Guided Token Search (GGTS)}, which uses scene geometry to guide two decisions: \textit{where to edit} and \textit{what to fill}. 
Geometric attribution determines \textit{where to edit} by selecting the token positions most responsible for the current violation while preserving unaffected parts of the trajectory. To decide \textit{what to fill}, the MGT proposes alternative motion tokens within the selected repair region, which are first screened by a geometric score and then verified through exact trajectory evaluation before an improving edit is finalised.
This targeted, coarse-to-exact search avoids unguided perturbations and exhaustive codebook scans; instead, a small number of targeted discrete replacements induce large, coordinated changes in motion structure, enabling exploration of alternative routes beyond the current planning basin.


We validate MGMP on diverse tasks and environments, showing that it generalizes across unseen layouts, additional obstacles and unseen geometries, scales to 14-DoF dual-arm planning, and transfers to real-world Baxter tasks.
We also introduce two diagnostic benchmarks: \textit{Ring Maze}, which requires substantial route changes to escape local planning basins, and \textit{Controlled Route Invalidation}, which fixes the same invalidated trajectory across methods to isolate repair from initial-generation quality. 
MGMP achieves 96\% success on Ring Maze and 82\% repair success on Controlled Route Invalidation, exceeding the strongest external baselines by 23 and 25 percentage points, respectively. 

In summary, this paper makes the following three contributions:
\begin{itemize}



\item We introduce MGMP, which reframes generative motion planning by extending the learned trajectory prior beyond efficient parallel generation to test-time structural repair.


\item We propose \textit{Geometry-Guided Token Search (GGTS)}, which turns test-time refinement into an efficient, targeted search over discrete motion alternatives, using scene geometry to determine where to edit and which candidates to evaluate, enabling structural repair when local continuous refinement is insufficient.

\item We introduce \textit{Ring Maze} and \textit{Controlled Route Invalidation} to evaluate basin escape and isolate structural repair from initial-generation quality, and validate MGMP across unseen obstacle layouts, additional obstacles, unseen geometries, and real-world Baxter tasks.

\end{itemize}

\begin{figure}[t]
    \centering
    \includegraphics[width=1\linewidth]{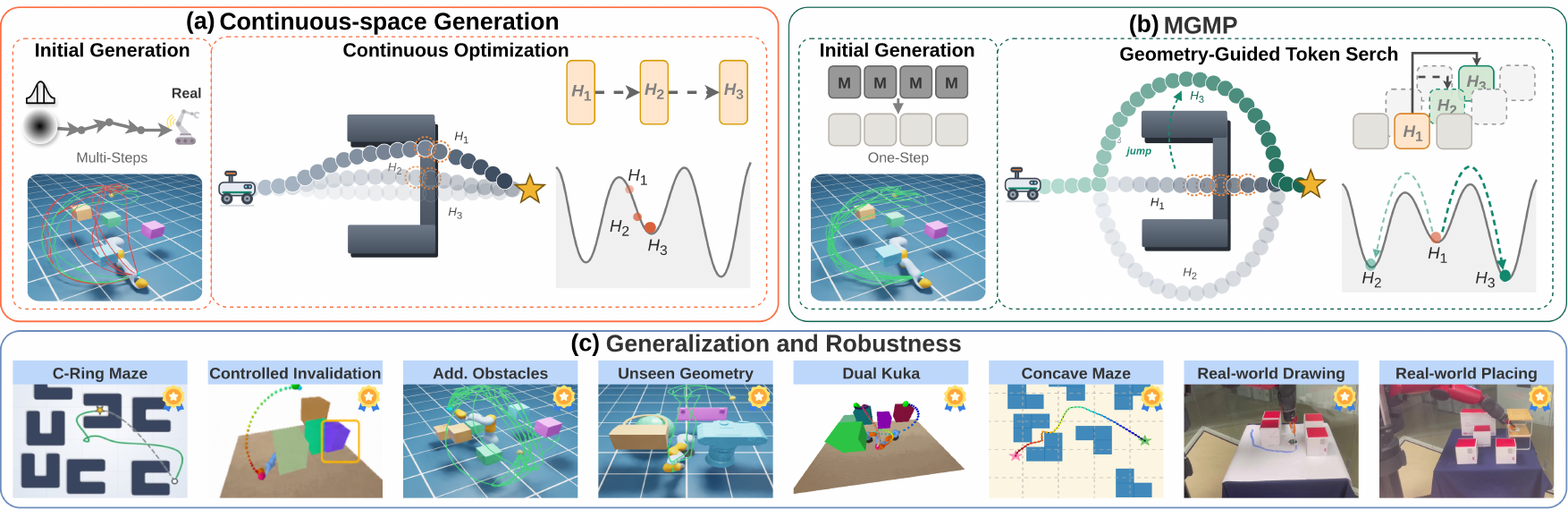}

    \vspace{-6pt}
    \caption{Overview of \textit{MGMP}.  We replace continuous trajectory generation and refinement with discrete masked generation and Geometry-Guided Token Search, enabling efficient planning and structural route repair. We evaluate it across challenging simulated and real-world settings, including basin escape, unseen geometries, additional obstacles, dual-arm planning, and Baxter tasks.
}

    \label{fig:main}
    \vspace{-4mm}
\end{figure}

\section{Related Work}
\label{gen_inst}

\vspace{-1mm}
\paragraph{Classical and learning-based motion planning.}
Classical planners either explore configuration space or locally refine a
trajectory. Sampling-based methods
~\citep{KavrakiSLO96,lavalle1998rapidly, karaman2011sampling,gammell2015batch} can discover alternative routes but often
require extensive sampling and collision checking in cluttered or
high-dimensional spaces
~\citep{orthey2023sampling,KavrakiSLO96,hauser2015lazy}.
Trajectory optimizers
~\citep{ratliff2009chomp,zucker2013chomp,kalakrishnan2011stomp,
schulman2013finding,mukadam2016gaussian} efficiently refine collision and
smoothness objectives, but remain sensitive to initialization and local planning
basins~\citep{pan2014predicting,schulman2014motion}.
This motivates learning reusable structure across planning problems
~\citep{qureshi2020motion,fishman2023motion,das2024motion}. Learning-based planners reduce online planning cost by learning sampling
distributions, predictors, policies, or latent representations
~\citep{qureshi2020motion,qureshi2018deeply,yu2021reducing,
fishman2023motion,yamada2023leveraging}.
Generative planners further model multimodal trajectory distributions, with
diffusion-based methods producing diverse candidates through iterative denoising
~\citep{janner2022planning,carvalho2023motion,luopotential}.
Because generated trajectories may violate scene constraints, they commonly
require guidance, correction, or trajectory optimization
~\citep{carvalho2023motion,luopotential,seo2025presto,
yang2026safeflowmatcher}.
Such continuous refinement is effective for local correction but typically cannot move from an infeasible region of configuration space to a feasible region when this requires a substantial route change to escape a local optimum.

\vspace{-2mm}

\vspace{-2mm}
\paragraph{Discrete representations and masked generative models.}
Discrete latent representations have been used to compress actions and
trajectories for planning and learned sampling
~\citep{jiang2022efficient,johnson2023learning}.
Masked generative models instead predict multiple discrete tokens in parallel
and iteratively revisit selected tokens over only a few refinement steps
~\citep{chang2022maskgit,chang2023muse}.
This paradigm has been applied to trajectory modeling, model-predictive control,
human motion generation, and robotic imitation learning
~\citep{wu2023masked,wen2025m,pinyoanuntapong2024mmm,
guo2024momask,zhuang2025masked}.
We extend masked generation to collision-constrained robot motion planning,
using learned motion tokens for efficient parallel trajectory generation.

\vspace{-3mm}
\paragraph{Test-time trajectory refinement and repair.}
Generated trajectories are commonly adapted to scene-specific constraints through
gradient-based guidance or trajectory optimization
~\citep{ratliff2009chomp,kalakrishnan2011stomp,schulman2014motion,
carvalho2023motion,luopotential,yang2026safeflowmatcher}.
These methods are effective for local geometric correction, but remain sensitive to initialization when feasibility requires a global route change~\citep{pan2014predicting,schulman2014motion}.
Recent masked generative methods revisit selected tokens during inference, but their update decisions are designed for generic generative quality rather than scene-specific geometric constraint violations~\citep{peng2025path,liu2025think}. 
Beyer and Karaman~\citet{beyer2026motion} instead search compressed trajectory tokens under test-time objectives, but use fixed coarse-to-fine traversal and candidate evaluation rather than a learned conditional prior to focus token proposals or geometric failure signals to target repair.
MGMP instead turns the learned trajectory prior itself into a geometry-guided, targeted search space for efficient structural repair over prior-supported route alternatives.


\section{Methodology}
\label{sec:method}

\begin{figure}[t]
    \centering
    \includegraphics[width=1\linewidth]{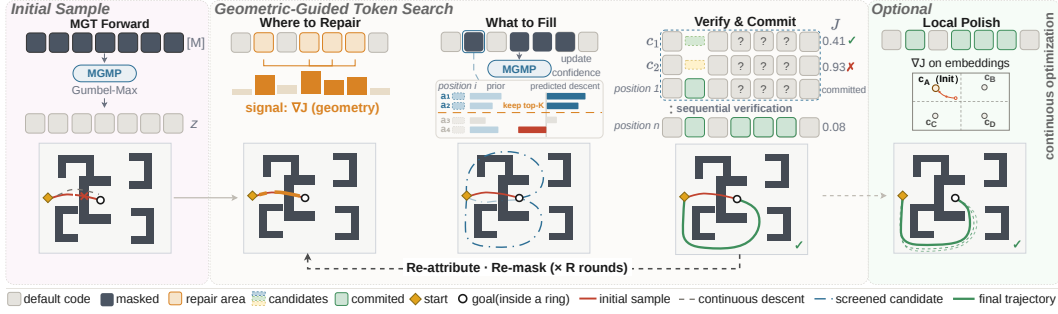}
    
    \vspace{-6pt}
    \caption{Overview of GGTS. We use geometric attribution to select edit positions, first-order scores to screen token replacements, and sequential verification to accept improving candidates, followed by optional continuous polish.
    }
    \label{fig:pipeline}
    \vspace{-6mm}
\end{figure}

\vspace{-2mm}
We formulate motion planning as conditional trajectory generation. Given a start state $c_s \in \mathbb{R}^{d}$, a goal state $c_g \in \mathbb{R}^{d}$, and environment constraints $w$, the planner is required to generate feasible motion trajectories $\left \{x_{t} \right \}_{t=0}^{T-1}$. Here $d$ is the dimensionality of the robot's state space, $x_{t}$ is its configuration at frame $t$, and $T$ is the horizon of the full trajectory. Sampled trajectories must satisfy $x_{0}=c_{s}$ and $x_{T-1}=c_{g}$, and the entire trajectory must be collision-free with respect to the environment.

Our method MGMP consists of a trajectory tokenizer and a conditional masked generative transformer (MGT). The tokenizer maps expert trajectories into discrete codes and reconstructs continuous trajectories (Sec.~\ref{sec:tokenizer}), while the MGT predicts masked codes conditioned on the start state, goal state, and environment (Sec.~\ref{sec:MGMP}). At test time, Geometry-Guided Token Search (GGTS) uses the learned predictor together with geometric feedback to repair the token trajectory (Fig.~\ref{fig:pipeline}; Sec.~3.3).


\vspace{-1mm}
\subsection{Representing Motion Configuration Space via Codebooks}
\label{sec:tokenizer}


To construct a compact discrete representation of feasible trajectories, we use a
VQ-VAE~\citep{van2017neural} with a learnable codebook $\mathcal{K}$.
The encoder maps $x\in\mathbb{R}^{T\times d}$ to latent representations
$\hat{\mathbf{e}}\in\mathbb{R}^{N\times d^\prime}$, which are quantized by
nearest-neighbor lookup to discrete token indices
$\mathbf{z}=(z_1,\ldots,z_N)$.
The codebook is
$\mathcal{K}=\{\mathbf{k}_m\}_{m=0}^{|\mathcal{K}|-1}$,
$\mathbf{k}_m\in\mathbb{R}^{d^\prime}$, with quantized embedding
$\mathbf{e}_i=\mathbf{k}_{z_i}$ at position $i$.
Each token summarizes approximately $T/N$ trajectory steps, and a symmetric
decoder reconstructs the continuous trajectory $\hat{x}$ from the quantized
embeddings $\mathbf{e}$.
Our VQ-VAE is trained with the reconstruction and commitment objective~\citep{van2017neural},
$
\mathcal{L}_{\text{VQ}}
=
\lambda_{\text{rec}}\|x-\hat{x}\|_1
+
\beta\|\hat{\mathbf{e}}-\mathrm{sg}[\mathbf{e}]\|_2^2,
$
together with a velocity loss 
to encourage temporal
continuity and, for robot trajectories, a forward-kinematics loss. After training, the tokenizer is frozen: the encoder maps expert
trajectories to discrete tokens for MGT training, while at inference time
predicted tokens are mapped to codebook embeddings and decoded into continuous
trajectories (See App.~\ref{app:network_architecture}).

\vspace{-1mm}
\subsection{Masked Generative Predictor over Codebooks}
\label{sec:MGMP}

We use a masked generative transformer (MGT) to model trajectories in the discrete space learned by the tokenizer. The MGT is an encoder-only transformer~\citep{vaswani2017attention} that predicts masked trajectory tokens conditioned on the start state $c_s$, goal state $c_g$, and obstacle information $w$. Start and goal states are projected into condition tokens, while obstacle centers are encoded by a lightweight transformer and injected into the MGT through AdaLN conditioning. During training, trajectories are first encoded into discrete tokens. A random subset is replaced by mask tokens, while some remaining tokens are randomly corrupted to alternative codebook indices. The MGT is trained with cross-entropy loss to recover the original token indices at the masked or corrupted positions, conditioned on the start, goal, and obstacle information. Details are provided in App.~\ref{app:network_architecture}.

\vspace{-1mm}
\subsection{Geometry-Guided Token Search}
\label{sec:sampling}

The trained mask transformer (Sec.~\ref{sec:MGMP}) provides a strong prior over trajectory tokens.
Given the planning condition $\mathbf{c}=(c_s,c_g,w)$, we obtain an initial trajectory through a single forward pass of the MGT, based on probability confidence; however, a high-confidence token can still produce a collision after decoding. We therefore formulate test-time refinement as a \textit{constraint-guided search in the learned discrete trajectory space}.
Following the plan-and-denoise view of masked generation~\citep{peng2025path,liu2025think}, each refinement round makes two decisions: which positions to re-mask (Sec.~\ref{sec:remask}) and what codes to put there (Sec.~\ref{sec:search}). Rather than relying solely on predictive uncertainty, we use an explicit geometric objective $J$ to guide both decisions, while the MGT prior preserves learned motion structure and smoothness.
At each editable position, the conditional MGT restricts the search to motion codes with high probability under the learned trajectory distribution, while geometry determines which positions and replacements are worth evaluating. This retains the ability to make larger discrete changes in trajectory structure while explicitly accounting for geometric constraints. The task-specific form of $J$ is detailed in App.~\ref{sec:task_scenarios}.

\subsubsection{Where to Edit: Geometry-Guided Adaptive Remasking 
}
\label{sec:remask}

The first MGT forward pass defines the token distribution $p_\theta(z_i\mid\mathbf{c})$ at each position. We initialize each token $z_i$ using Gumbel-Max sampling.
We employ the \textit{Gauss-Southwell rule} to select the regions requiring refinement---choosing those tokens where the magnitude of the gradient is greatest. 
Specifically, we backpropagate $J$ through the decoder to the current token embeddings $\mathbf{e}_i$ and define
\begin{equation}
    s_i =
    \left\|
        \frac{\partial J}{\partial \mathbf{e}_i}
    \right\|_2
    \left(1+\lambda_{\mathrm{conf}}(1-c_i)\right),
    \label{eq:mask_score}
\end{equation}
where $c_i$ is the MGT confidence of token $i$ and $\lambda_{\text{conf}}$ is a scale factor. The geometric gradient provides the primary signal, while confidence provides a secondary modulation. Thus, a high-confidence token can still be replaced when it strongly contributes to a geometric violation.

We rank the token positions by $s_i$ and re-mask the smallest top-ranked set whose cumulative score accounts for a fraction $\rho$ of the total score. Token positions associated with detected collision timesteps are additionally included in the re-masking set. The mask size thus adapts to the failure: a shallow contact considers replacing only a few tokens, while a trajectory on the wrong side of an obstacle exposes a larger set of trajectory coordinates, which enables substantial restructuring. Such changes can enable substantial route changes (see Sec.~\ref{sec:kuka_invalidation}).

\subsubsection{What to Fill: Candidate Proposal, Screening, and Verification}
\label{sec:search}

\paragraph{Learned proposals and first-order screening.}

After selecting the positions to re-mask, we perform one masked MGT forward pass. At each selected position $i$, we retain the top $M$ codes under the MGT distribution as candidate replacements, restricting the search to discrete alternatives supported by the learned trajectory prior. Evaluating every proposal through full decoding is expensive, so we first screen the candidates using a Gumbel-Softmax relaxation. Let $\tilde p_{i,j}$ denote the Gumbel-Softmax weight of candidate $j$ among the $M$ proposals after adding Gumbel noise, and let $\mathbf{e}_{i,j}$ denote its codebook embedding. At each selected position, we form a probability-weighted relaxed embedding while keeping all other positions at their current hard embeddings.
Then each candidate is scored using a first-order approximation of the geometric objective:
\begin{equation}
    r_{i,j}
    =
    -\,\tilde p_{i,j}
        \frac{\partial J}{\partial \bar{\mathbf e}_i} \cdot
        (\mathbf{e}_{i,j}-\bar{\mathbf e}_i),
    \qquad
    \bar{\mathbf e}_i
    =
    \sum_{j=1}^{M}
    \tilde p_{i,j}\mathbf{e}_{i,j}
    \label{eq:candidate_screen}
\end{equation}
where $\bar{\mathbf e}_i$ is the relaxed embedding at position $i$, and $r_{i,j}$ is the first-order screening score for candidate $j$. The Gumbel-Softmax weight favors replacements supported by the learned prior, while the inner product favors candidates aligned with a local descent direction of $J$. We retain the top-$k$ candidates according to $r_{i,j}$ for exact verification. While the gradient in Eq.~\ref{eq:mask_score} determines \emph{where} the trajectory is edited, the gradient in Eq.~\ref{eq:candidate_screen} ranks \emph{what} tokens are worth testing at those positions.

\paragraph{Verifier-guided sequential refinement.}

The first-order score is used only for screening, since it does not capture decoder nonlinearities or interactions between token replacements. We therefore verify shortlisted candidates using hard-token trajectories. Selected positions are visited in decreasing attribution order, and each candidate is substituted into the current token sequence, decoded, and evaluated with the exact objective $J$. 
For compactness, we write $\mathcal{D}(\mathbf{z})$ to denote decoding the codebook embeddings corresponding to token sequence $\mathbf{z}$. 
The best replacement is committed only if it improves the current solution:
\begin{equation}
z_i \leftarrow
\begin{cases}
\displaystyle
\operatorname*{arg\,min}_{z_i'\in\mathcal{N}_i(\mathbf{z})}
J\!\left(\mathcal{D}\!\left(\mathbf{z}^{(i\leftarrow z_i')}\right)\right),
&
\displaystyle
\min_{z_i'\in\mathcal{N}_i(\mathbf{z})}
J\!\left(\mathcal{D}\!\left(\mathbf{z}^{(i\leftarrow z_i')}\right)\right)
<
J\!\left(\mathcal{D}(\mathbf{z})\right),\\[8pt]
z_i, & \text{otherwise.}
\end{cases}
\label{eq:sequential_commit}
\end{equation}
where $\mathbf{z}^{(i\leftarrow z_i')}$ denotes the full sequence with position $i$ replaced by $z_i'$, and $\mathcal{N}_i(\mathbf{z})$ is the shortlist at that position. Accepted edits are applied immediately, so subsequent positions are evaluated against the updated trajectory. If none of the shortlisted candidates at the
current position improves the solution, the corresponding position--token pairs are recorded as failed candidates and excluded from subsequent evaluation.

\subsubsection{Iterative Refinement}

At each round, we recompute how strongly each token contributes to the current geometric violation, producing a new re-masking set and MGT proposal pool. Candidate screening and sequential verification are then performed against the latest committed trajectory, so later rounds focus on the remaining violations. Refinement stops once the trajectory is feasible or the maximum number of rounds is reached. We then optionally apply only a few gradient steps of continuous optimization to the final token embeddings. Thus, discrete token replacement provides structural repair, while continuous polishing performs only local correction. Refinement is summarized in App.~\ref{app:algorithm}.


\section{Experiments}
\label{sec:experiment}

\paragraph{Evaluation setup.}
We organize the evaluation from controlled analysis of structural repair to broader generalization and real-world execution. 
\textit{Ring Maze} tests whether MGMP can escape local planning basins that require substantial route changes (Sec.~\ref{sec:ring_maze}), while \textit{Controlled Route Invalidation on Kuka} isolates GGTS-based repair from initial-generation quality by initializing all methods from the same invalidated trajectory (Sec.~\ref{sec:kuka_invalidation}).
We then evaluate end-to-end planning on Maze2D, 7-DoF Kuka, and 14-DoF DualKuka under unseen obstacle layouts, additional obstacles, and unseen obstacle geometries (Sec.~\ref{sec:generalization}), followed by ablations (Sec.~\ref{sec:ablation}) and real-world Baxter experiments (Sec.~\ref{sec:real}).
For the end-to-end benchmarks, we evaluate 1,000 planning problems across 50 unseen environments per task.
Full environment construction, training-data generation, and evaluation protocols are provided in App.~\ref{app:env_details}.



\vspace{-3mm}
\paragraph{Baselines and variants.}
We compare MGMP with eleven classical and learning-based planners: RRT*~\citep{karaman2011sampling}, P-RRT*~\citep{qureshi2016potential}, BIT*~\citep{gammell2015batch}, RMP~\citep{ratliff2018riemannian}, CHOMP~\citep{ratliff2009chomp}, TrajOpt~\citep{schulman2014motion}, STOMP~\citep{kalakrishnan2011stomp}, M$\pi$Nets~\citep{fishman2023motion}, MPD~\citep{carvalho2023motion}, PBD~\citep{luopotential}, and SafeFlowMatcher~\citep{yang2026safeflowmatcher}.
We further compare \emph{MGT+EmbOpt (GD-30)}, which replaces GGTS with 30 steps of continuous embedding optimization, and \emph{MGMP w/o Polish}, which removes the final five-step continuous polishing stage.
See App.~\ref{app:baseline_details} for details.

\vspace{-3mm}
\paragraph{Metrics.}
 
For Ring Maze and end-to-end planning, we report \emph{success rate (SR)}, \emph{valid-sample fraction (VSF)}, \emph{planning time (PT)}, and \emph{running time (RT)}.
Each generative planner produces 20 candidates per problem; SR measures whether at least one is valid, while VSF is the fraction of valid candidates.
PT is the time to obtain a feasible solution, whereas RT measures the wall-clock cost of generating and refining the full candidate set. Details are provided in App.\ref{app:plan_time}. 
For \emph{Controlled Route Invalidation}, we report \emph{repair success rate (RS)} defined as the fraction of invalidated trajectories repaired into valid solutions.
Among successful repairs, we additionally report the average and maximum joint-space deviation from the shared initial trajectory, together with the longest contiguous changed span. Full definitions are provided in App.~\ref{app:control_kuka}.

\vspace{-1mm}
\subsection{Escaping Local Planning Basins}
\label{sec:escape}


We first use \textit{Ring Maze} to test whether MGMP can escape local optima during planning that hard be resolved by local adjustments, and then use \textit{Controlled Route Invalidation on Kuka} to isolate the contribution of GGTS-based trajectory repair from differences in initial trajectory generation.



\vspace{-2mm}
\subsubsection{Natural Basin Traps in Ring Maze.}
\label{sec:ring_maze}

We construct a compact 2-D Ring Maze benchmark with seven near-enclosing C-shaped obstacles, a standard geometry for inducing local planning minima. Their dense arrangement creates route-level traps, where escaping one enclosure may still require further route changes to reach the goal. Consequently, successful planning can require substantial route changes across the environment rather than local waypoint adjustments around individual collisions. To ensure sufficient coverage of such cases, we include start–goal pairs with states inside the C-shaped enclosures. We evaluate 4,000 problems across 100 unseen environments, of which $70\%$ require escaping an enclosure. Environment, dataset, and implementation details are provided in Apps.~\ref{app:env_details} and~\ref{sec:ring_impl}.

MGMP achieves the highest success rate (96\%) among the evaluated refinement-based methods (Table~\ref{tab:basin_escape}). The improvement is particularly pronounced in valid-sample fraction, where MGMP reaches 67\%, compared with 29\% for MGT + EmbOpt and 23\% for the best classical trajectory optimizer. MGMP w/o Polish already achieves 91\% success and 64\% valid-sample fraction, showing that GGTS provides most of the basin-escape capability, while the final polishing stage further improves feasibility. See Fig.~\ref{fig:ringmaze_kuka}(a) for representative Ring Maze trajectories.
MGMP achieves these gains with only 0.08s runtime.
Further analysis in App.~\ref{app:ring_analysis} shows that increasing the local-refinement budget does not close this gap: CHOMP changes little from 10--50 optimization iterations, while MPD improves only up to three guidance steps before degrading. As collision severity increases, PBD's repair rate drops sharply, whereas MGMP remains relatively stable, showing that GGTS is more capable of repairing deeply invalid trajectories that require substantial route changes.

\begin{table*}[t]
\centering
\caption{
Results on Ring Maze and Controlled Route Invalidation on Kuka.
SR, VSF, RT, and RS denote success rate, valid-sample fraction, running time,
and repair success rate, respectively.
For Controlled Route Invalidation, trajectory-deviation statistics are reported
over successful repairs.
}
\label{tab:basin_escape}
\resizebox{0.85\textwidth}{!}{%
\begin{tabular}{@{}lccccccc@{}}
\toprule
\textbf{Methods}
& \multicolumn{3}{c}{\textbf{Ring Maze}}
& \multicolumn{4}{c}{\textbf{Controlled Route Invalidation on Kuka}} \\
\cmidrule(lr){2-4}
\cmidrule(l){5-8}
& SR$\uparrow$
& VSF$\uparrow$
& RT$\downarrow$
& RS$\uparrow$
& Avg. Dev.
& Max. Dev.
& Changed Span \\
\midrule

CHOMP
& $0.73$\std{0.09}
& $0.23$\std{0.10}
& $0.56$\std{0.14}
& $0.46$\std{0.15}
& $0.82$\std{0.19}
& $1.30$\std{0.32}
& $31.20$\std{9.40}
 \\

TrajOpt
& $0.66$\std{0.04}
& $0.22$\std{0.03}
& $2.64$\std{0.01}
& $0.53$\std{0.14}
& $0.88$\std{0.20}
& $1.61$\std{0.38}
& $34.80$\std{10.10}
 \\

STOMP
& $0.23$\std{0.01}
& $0.05$\std{0.03}
& $0.45$\std{0.07}
& $0.57$\std{0.15}
& $0.96$\std{0.22}
& $1.66$\std{0.35}
& $39.60$\std{10.80}
\\

\midrule

MPD
& $0.14$\std{0.06}
& $0.02$\std{0.02}
& $0.21$\std{0.06}
& $0.38$\std{0.13}
& $0.83$\std{0.20}
& $1.43$\std{0.35}
& $31.90$\std{9.60}
 \\

PBD
& $0.67$\std{0.10}
& $0.26$\std{0.09}
&  $0.16$\std{0.05}
& $0.44$\std{0.14}
& $0.93$\std{0.22}
& $1.65$\std{0.39}
& $37.20$\std{10.30}

 \\

SafeFlowMatcher
& $0.10$\std{0.05}
& $0.08$\std{0.06}
& $2.33$\std{0.32}
& $0.31$\std{0.14}
& $0.91$\std{0.21}
& $1.70$\std{0.42}
& $37.10$\std{10.30}
 \\

\midrule

MGT+EmbOpt30
& $0.75$\std{0.09}
& $0.29$\std{0.11}
& $0.08$\std{0.02}
& $0.60$\std{0.15}
& $1.02$\std{0.24}
& $1.71$\std{0.36}
& $45.20$\std{11.10}
 \\

MGMP w/o Polish
& $0.91$\std{0.06}
& $0.64$\std{0.13}
& \cellcolor{gray!15}$\mathbf{0.07}$\std{0.01}
& $0.76$\std{0.13}
& $1.42$\std{0.28}
& $2.65$\std{0.50}
& $73.40$\std{11.60}
 \\

MGMP (Ours)
& \cellcolor{gray!15}$\mathbf{0.96}$\std{0.04}
& \cellcolor{gray!15}$\mathbf{0.67}$\std{0.12}
& $0.08$\std{0.02}
& \cellcolor{gray!15}$\mathbf{0.82}$\std{0.12}
& $1.38$\std{0.27}
& $2.40$\std{0.44}
& $71.70$\std{11.20}
 \\

\bottomrule
\end{tabular}%
}
\end{table*}

\begin{figure}[t]
    \centering
    \includegraphics[width=1\linewidth]{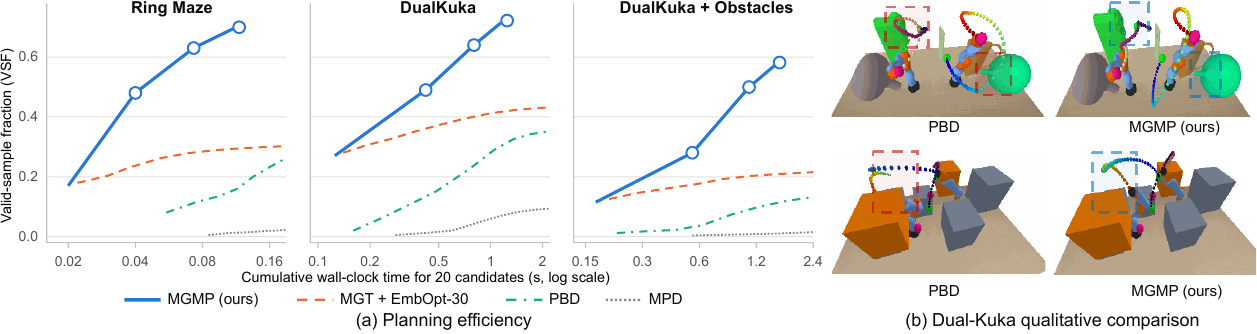}
    
    \vspace{-6pt}
\caption{
Planning efficiency and qualitative comparison.
(a) Valid-sample fraction versus wall-clock time across three settings.
(b) Dual-Kuka results under unseen obstacle geometries (top) and additional obstacles (bottom); PBD collides (red), while MGMP is collision-free (blue).
}
    
    \label{fig:efficiency_qua}
    \vspace{-5mm}
\end{figure}

\vspace{-2mm}
\subsubsection{Controlled Route Invalidation on Kuka}
\label{sec:kuka_invalidation}

\begin{figure}[t]
    \centering
    \includegraphics[width=0.8\linewidth]{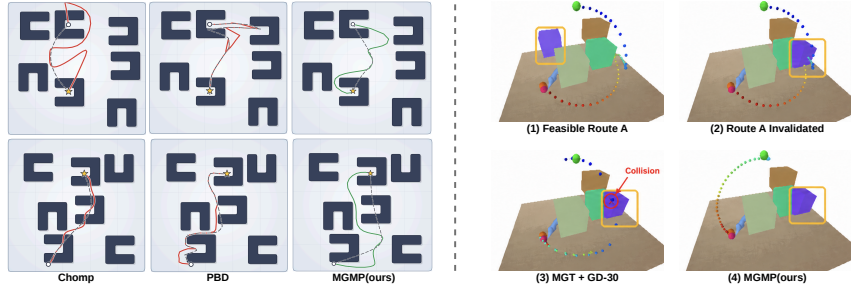}
    
    \vspace{-6pt}
\caption{(a) Qualitative comparison on Ring Maze.
Top and bottom rows show two planning problems. Dashed/solid curves denote initial/refined trajectories, and green/red denote collision-free/colliding solutions.
(b) Example of Controlled Route Invalidation on Kuka.
}
    
    \label{fig:ringmaze_kuka}
    \vspace{-3mm}
\end{figure}




To isolate structural repair capability from initial-generation quality, we evaluate 500 invalidation problems across 50 unseen 7-DoF Kuka environments. For each problem, we first obtain a verified feasible Route A and pass it through the trajectory tokenizer to obtain a reconstructed route. We verify that this reconstructed route remains feasible in the original scene, and then relocate one obstacle to invalidate it while preserving a verified alternative Route B that is substantially separated from the invalidated Route A in joint space.
All methods are initialized from the same reconstructed route under the same modified constraints and directly apply only their respective refinement procedures, without generating a new initial trajectory. The modified scene therefore requires a structural change to the shared initial route rather than a local adjustment around the newly introduced collision. A representative example is shown in Fig.~\ref{fig:ringmaze_kuka} (b). Initialization details are provided in App.~\ref{app:control_kuka}.

As shown in Table~\ref{tab:basin_escape}, MGMP achieves 82\% repair success, outperforming MGT+EmbOpt (60\%) and the strongest classical optimizer, STOMP (57\%). MGMP w/o Polish already reaches 76\%, showing that GGTS provides most of the improvement. 
Among MGMP's successful repairs, the longest changed span
covers 71.7\% of the trajectory, indicating that GGTS can modify extended portions of the route rather than only the immediate collision region. The similar repair
magnitude of MGMP w/o Polish and full MGMP, together with the increase in repair
success from 76\% to 82\% after polishing, supports complementary roles for the
two stages: GGTS provides the large-scale trajectory repair, while polishing
resolves residual local infeasibilities.

\vspace{-1mm}
\subsection{Generalization and Robustness}
\label{sec:generalization}

\vspace{-1mm}
\paragraph{Generalization to Unseen Obstacle Layouts}
\label{sec:base_env}
We first evaluate generalization to unseen obstacle layouts, where obstacle types and counts match the training environments, but obstacle placements and start-goal pairs are unseen (Table~\ref{tab:base_all}; implementation details in App.~\ref{app:impl_unseen_layouts}).
On Maze2D, MGMP achieves $100\%$ success on both tasks, matching BIT* with lower planning time, while attaining the highest valid-sample fractions among learning-based planners. On 7-DoF Kuka, MGMP maintains a success rate comparable to BIT* while reducing planning time from $0.39$\,s to $0.13$\,s. It also improves the valid-sample fraction over PBD from $0.65$ to $0.87$. The advantage is more pronounced on the higher-dimensional 14-DoF DualKuka task: MGMP achieves a success rate of 0.96, compared with 0.85 for BIT*, 0.81 for CHOMP, and 0.79 for PBD, while reducing planning time from 2.01\,s for BIT* to 0.39\,s and achieving a valid-sample fraction of 0.70, substantially higher than PBD's 0.35. Overall, MGMP maintains high success rates on unseen layouts while producing more valid trajectories with lower runtime. Detailed results are provided in App.~\ref{app:per_env_unseen_layouts}.

\begin{table*}[t]
\centering
\caption{Results on unseen obstacle layouts. SR, VSF, and PT denote success rate, valid-sample fraction, and planning time in seconds. Results are reported as mean $\pm$ standard deviation.}

\label{tab:base_all}
\resizebox{\textwidth}{!}{%
\setlength{\tabcolsep}{2pt}
\begin{tabular}{@{}lcccccccccccc@{}}
\toprule
\textbf{Methods} 
& \multicolumn{3}{c}{\textbf{Maze2D-6}}
& \multicolumn{3}{c}{\textbf{Concave Maze2D}}
& \multicolumn{3}{c}{\textbf{7-DoF Kuka}}
& \multicolumn{3}{c}{\textbf{14-DoF DualKuka}} \\
\cmidrule(lr){2-4} 
\cmidrule(lr){5-7} 
\cmidrule(lr){8-10} 
\cmidrule(l){11-13}
& SR$\uparrow$ & VSF$\uparrow$ & PT$\downarrow$
& SR$\uparrow$ & VSF$\uparrow$ & PT$\downarrow$ 
& SR$\uparrow$ & VSF$\uparrow$ & PT$\downarrow$ 
& SR$\uparrow$ & VSF$\uparrow$ & PT$\downarrow$  \\
\midrule
RMP 
& $0.44$\std{0.13} & -- & $3.09$\std{0.59}
& $0.38$\std{0.13} & -- & $3.42$\std{0.64}
& $0.81$\std{0.14} & -- & $1.24$\std{0.72}
& $0.65$\std{0.18} & -- & $2.25$\std{0.88} \\

RRT* 
& $0.99$\std{0.01} & -- & $0.54$\std{0.13}
& $0.98$\std{0.03} & -- & $1.24$\std{0.22}
& $0.64$\std{0.17} & -- & $2.31$\std{0.78}
& $0.44$\std{0.17} & -- & $3.49$\std{0.64} \\

P-RRT*
& $0.99$\std{0.02} & -- & $0.86$\std{0.18}
& $0.96$\std{0.05} & -- & $1.60$\std{0.27}
& $0.63$\std{0.18} & -- & $2.46$\std{0.80}
& $0.42$\std{0.17} & -- & $3.58$\std{0.64} \\

BIT* 
& \cellcolor{gray!15}$\mathbf{1.00}$\std{0.01} & -- & $0.15$\std{0.02}
& \cellcolor{gray!15}$\mathbf{1.00}$\std{0.01} & -- & $0.29$\std{0.04}
& \cellcolor{gray!15}$\mathbf{0.97}$\std{0.02} & -- & $0.39$\std{0.25}
& $0.85$\std{0.14} & -- & $2.01$\std{0.76} \\
\midrule

CHOMP
& $0.99$\std{0.01} & $0.64$\std{0.07} & $0.35$\std{0.03}
& $0.99$\std{0.01} & $0.55$\std{0.02} & $0.24$\std{0.08}
& $0.89$\std{0.04} & $0.62$\std{0.02} & $1.50$\std{0.08}
& $0.81$\std{0.05} & $0.49$\std{0.04} & $1.74$\std{0.12} \\

TrajOpt
& $0.99$\std{0.01} & $0.87$\std{0.01} & $0.47$\std{0.02}
& $0.99$\std{0.01} & $0.72$\std{0.07} & $1.91$\std{0.02}
& $0.67$\std{0.03} & $0.25$\std{0.06} & $2.41$\std{0.01}
& $0.67$\std{0.08} & $0.25$\std{0.06} & $3.95$\std{0.23} \\

STOMP
& $0.91$\std{0.05} & $0.34$\std{0.04} & $0.47$\std{0.08}
& $0.85$\std{0.07} & $0.27$\std{0.06} & $0.35$\std{0.08}
& $0.78$\std{0.05} & $0.27$\std{0.02} & $1.23$\std{0.04}
& $0.64$\std{0.09} & $0.14$\std{0.08} & $1.88$\std{0.16} \\

\midrule

M$\pi$Nets 
& $0.98$\std{0.01} & -- & $0.13$\std{0.09}
& $0.94$\std{0.02} & -- & $0.14$\std{0.02}
& $0.85$\std{0.08} & -- & $0.23$\std{0.01}
& $0.61$\std{0.01} & -- & $0.74$\std{0.03} \\

MPD 
& $0.83$\std{0.10} & $0.31$\std{0.04} & $0.14$\std{0.03}
& $0.59$\std{0.11} & $0.30$\std{0.04} & $0.27$\std{0.08}
& $0.62$\std{0.28} & $0.35$\std{0.12} & $0.35$\std{0.03}
& $0.28$\std{0.21} & $0.09$\std{0.09} & $0.73$\std{0.09} \\

PBD
& \cellcolor{gray!15}$\mathbf{1.00}$\std{0.01} & $0.92$\std{0.04} & $0.15$\std{0.06}
& $0.98$\std{0.01} & $0.88$\std{0.07} & $0.15$\std{0.09}
& $0.93$\std{0.01} & $0.65$\std{0.17} & $0.15$\std{0.09}
& $0.79$\std{0.15} & $0.35$\std{0.17} & $0.42$\std{0.26} \\

S.F.Matcher
& $0.50$\std{0.03} & $0.29$\std{0.05} & $1.86$\std{0.12}
& $0.45$\std{0.01} & $0.32$\std{0.01} & $2.00$\std{0.16}
& $0.45$\std{0.06} & $0.28$\std{0.03} & $2.47$\std{0.02}
& $0.29$\std{0.04} & $0.13$\std{0.01} & $2.86$\std{0.05} \\

MGMP (Ours)\hspace{-2pt}
& \cellcolor{gray!15}$\mathbf{1.00}$\std{0.01} & \cellcolor{gray!15}$\mathbf{0.98}$\std{0.01} & \cellcolor{gray!15}$\mathbf{0.03}$\std{0.01}
& \cellcolor{gray!15}$\mathbf{1.00}$\std{0.01} & \cellcolor{gray!15}$\mathbf{0.95}$\std{0.04} & \cellcolor{gray!15}$\mathbf{0.07}$\std{0.01}
& \cellcolor{gray!15}$\mathbf{0.97}$\std{0.07} & \cellcolor{gray!15}$\mathbf{0.87}$\std{0.14} & \cellcolor{gray!15}$\mathbf{0.13}$\std{0.07}
& \cellcolor{gray!15}$\mathbf{0.96}$\std{0.06} & \cellcolor{gray!15}$\mathbf{0.70}$\std{0.13} & \cellcolor{gray!15}$\mathbf{0.39}$\std{0.12} \\
\bottomrule
\end{tabular}%
}
\vspace{-5mm}
\end{table*}

\vspace{-2mm}
\paragraph{Generalization to Additional-Obstacle Environments}
\label{sec:comp_env}



We next evaluate generalization to environments where scenes contain more obstacles than those observed during training. Additional setup details are provided in App.~\ref{sec:Additional-Obstacle Environments Set up}.
%
%
As shown in Table~\ref{tab:generalization_env}, MGMP remains effective as number of obstacles increases. On 7-DoF Kuka, MGMP slightly improves success over BIT* ($0.96$ vs.~$0.95$) while reducing planning time from $0.73$ s to $0.23$ s, and increases the valid-sample fraction over CHOMP from $0.52$ to $0.86$. The advantage is larger on 14-DoF DualKuka, where MGMP improves success over BIT* from $0.63$ to $0.84$ and reduces planning time from $3.44$ s to $0.90$ s, while increasing the valid-sample fraction over CHOMP from $0.26$ to $0.54$. 
Taken together, these results indicate that MGMP is less sensitive to increased test-time constraint density, particularly in the higher-dimensional DualKuka setting, as further illustrated qualitatively in Fig.~\ref{fig:efficiency_qua}(b) and App.~\ref{sec:Generalization to Additional-Obstacle Environments app}.

\paragraph{Generalization to Unseen Obstacle Geometries}
\label{sec:changing environments}

We further evaluate generalization to unseen obstacle geometries by replacing the box obstacles used during training with ShapeNet~\citep{shapenet} object meshes. We use ten everyday object categories and randomly sample four objects for each Kuka environment and five for each DualKuka environment. DeepSDF~\citep{park2019deepsdf} provides differentiable collision costs for the unseen meshes during refinement. Further details are provided in App.~\ref{sec:Unseen Obstacle Geometries Environments Set up}. As shown in Table~\ref{tab:generalization_env}, On 7-DoF Kuka, it achieves $0.66$ valid-sample fraction, compared with $0.47$ for PBD. On 14-DoF DualKuka, MGMP achieves the highest success rate among all methods ($0.87$), compared with $0.83$ for BIT* and $0.75$ for PBD, while increasing the valid-sample fraction over CHOMP from $0.34$ to $0.57$. 
These results show that MGMP remains effective under substantial changes in obstacle geometry, as further illustrated in Fig.~\ref{fig:efficiency_qua}(b) and App.~\ref{sec:Generalization to Unseen Obstacle Geometries app}.

\begin{table*}[t]
\centering
\caption{
Results on additional-obstacle and unseen-geometry environments.
SR, VSF, and PT denote success rate, valid-sample fraction, and planning time
in seconds, respectively.
}
\label{tab:generalization_env}
\vspace{-4pt}

\resizebox{0.95\textwidth}{!}{%
\setlength{\tabcolsep}{2.5pt}
\begin{tabular}{@{}lcccccccccccc@{}}
\toprule
\multirow{2}{*}{\textbf{Methods}}
& \multicolumn{3}{c}{\textbf{Kuka Add.-Obs.}}
& \multicolumn{3}{c}{\textbf{DualKuka Add.-Obs.}}
& \multicolumn{3}{c}{\textbf{Kuka Unseen Geo.}}
& \multicolumn{3}{c}{\textbf{DualKuka Unseen Geo.}} \\
\cmidrule(lr){2-4}
\cmidrule(lr){5-7}
\cmidrule(lr){8-10}
\cmidrule(lr){11-13}

& SR$\uparrow$ & VSF$\uparrow$ & PT$\downarrow$
& SR$\uparrow$ & VSF$\uparrow$ & PT$\downarrow$
& SR$\uparrow$ & VSF$\uparrow$ & PT$\downarrow$
& SR$\uparrow$ & VSF$\uparrow$ & PT$\downarrow$ \\
\midrule

RMP
& $0.77$\std{0.22} & -- & $1.40$\std{1.12}
& $0.48$\std{0.19} & -- & $3.05$\std{0.88}
& $0.77$\std{0.15} & -- & $1.47$\std{0.78}
& $0.65$\std{0.17} & -- & $2.21$\std{0.80} \\

RRT*
& $0.65$\std{0.23} & -- & $2.24$\std{1.03}
& $0.30$\std{0.16} & -- & $4.01$\std{0.56}
& $0.60$\std{0.15} & -- & $2.58$\std{0.73}
& $0.46$\std{0.19} & -- & $3.44$\std{0.74} \\

P-RRT*
& $0.66$\std{0.23} & -- & $2.38$\std{1.02}
& $0.29$\std{0.17} & -- & $4.14$\std{0.54}
& $0.57$\std{0.15} & -- & $2.84$\std{0.70}
& $0.43$\std{0.18} & -- & $3.58$\std{0.70} \\

BIT*
& $0.95$\std{0.10} & -- & $0.73$\std{0.62}
& $0.63$\std{0.21} & -- & $3.44$\std{0.81}
& \cellcolor{gray!15}$\mathbf{0.91}$\std{0.10} & -- & $0.99$\std{0.58}
& $0.83$\std{0.19} & -- & $1.88$\std{0.96} \\

\midrule

CHOMP
& $0.77$\std{0.02} & $0.52$\std{0.04} & $2.16$\std{0.05}
& $0.68$\std{0.07} & $0.26$\std{0.06} & $2.95$\std{0.01}
& $0.65$\std{0.02} & $0.33$\std{0.01} & $2.20$\std{0.06}
& $0.57$\std{0.08} & $0.34$\std{0.05} & $2.68$\std{0.08} \\

TrajOpt
& $0.68$\std{0.02} & $0.27$\std{0.01} & $2.35$\std{0.01}
& $0.60$\std{0.01} & $0.11$\std{0.03} & $3.80$\std{0.28}
& $0.63$\std{0.05} & $0.37$\std{0.02} & $2.35$\std{0.18}
& $0.74$\std{0.03} & $0.31$\std{0.07} & $3.34$\std{0.39} \\

STOMP
& $0.73$\std{0.01} & $0.28$\std{0.06} & $1.18$\std{0.06}
& $0.41$\std{0.09} & $0.07$\std{0.06} & $1.70$\std{0.19}
& $0.66$\std{0.02} & $0.23$\std{0.02} & $1.38$\std{0.15}
& $0.57$\std{0.01} & $0.18$\std{0.06} & $1.38$\std{0.15} \\

\midrule

M$\pi$Nets
& $0.52$\std{0.06} & -- & $0.42$\std{0.03}
& $0.29$\std{0.09} & -- & $1.97$\std{0.11}
& -- & -- & --
& -- & -- & -- \\

MPD
& $0.37$\std{0.27} & $0.08$\std{0.10} & $0.46$\std{0.04}
& $0.12$\std{0.11} & $0.01$\std{0.03} & $1.80$\std{0.35}
& $0.34$\std{0.11} & $0.14$\std{0.17} & $0.68$\std{0.02}
& $0.30$\std{0.04} & $0.11$\std{0.06} & $1.98$\std{0.09} \\

PBD
& $0.76$\std{0.17} & $0.50$\std{0.19} & $0.47$\std{0.08}
& $0.51$\std{0.22} & $0.12$\std{0.10} & $1.33$\std{0.19}
& $0.76$\std{0.19} & $0.47$\std{0.19} & $0.23$\std{0.02}
& $0.75$\std{0.19} & $0.36$\std{0.19} & $0.96$\std{0.08} \\

S.F.Matcher
& $0.38$\std{0.09} & $0.28$\std{0.03} & $2.46$\std{0.05}
& $0.20$\std{0.03} & $0.16$\std{0.02} & $1.87$\std{0.05}
& $0.31$\std{0.07} & $0.21$\std{0.01} & $2.47$\std{0.03}
& $0.33$\std{0.03} & $0.17$\std{0.02} & $2.92$\std{0.01} \\

MGMP (Ours)
& \cellcolor{gray!15}$\mathbf{0.96}$\std{0.05}
& \cellcolor{gray!15}$\mathbf{0.86}$\std{0.14}
& \cellcolor{gray!15}$\mathbf{0.23}$\std{0.12}
& \cellcolor{gray!15}$\mathbf{0.84}$\std{0.09}
& \cellcolor{gray!15}$\mathbf{0.54}$\std{0.16}
& \cellcolor{gray!15}$\mathbf{0.90}$\std{0.21}
& $0.90$\std{0.08}
& \cellcolor{gray!15}$\mathbf{0.66}$\std{0.13}
& \cellcolor{gray!15}$\mathbf{0.21}$\std{0.10}
& \cellcolor{gray!15}$\mathbf{0.87}$\std{0.09}
& \cellcolor{gray!15}$\mathbf{0.57}$\std{0.14}
& \cellcolor{gray!15}$\mathbf{0.72}$\std{0.18} \\

\bottomrule
\end{tabular}%
}

\vspace{-3mm}
\end{table*}


\vspace{-2mm}
\paragraph{Planning efficiency}

%

Figure~\ref{fig:efficiency_qua} (a) compares the quality--time trade-off of different refinement strategies. MGMP starts from a strong initial candidate set and improves rapidly with only a few GGTS rounds, increasing valid-sample fraction from 0.18 to 0.67 on Ring Maze and from 0.28 to 0.70 on DualKuka. The same trend holds under additional obstacles, where MGMP reaches about 0.54 valid-sample fraction, compared with roughly 0.21 for MGT+EmbOpt-30 and 0.14 for PBD. These results show that MGMP benefits from both strong discrete initialization and fast geometry-guided repair, reaching substantially higher-quality candidate sets within a small inference budget.

\subsection{Ablation Study}
\label{sec:ablation}

\begin{wraptable}{r}{0.50\columnwidth}
\centering

\caption{
Ablation results on Ring Maze and 7-DoF Kuka.
SR, VSF, and RT denote success rate, valid-sample fraction,
and runtime.
}
\label{tab:ggts_ablation}

\vspace{-0.4em}

\resizebox{\linewidth}{!}{
\begin{tabular}{lcccccc}
\toprule
\multirow{2}{*}{Variant}
&
\multicolumn{3}{c}{Ring Maze}
&
\multicolumn{3}{c}{7-DoF Kuka}
\\
\cmidrule(lr){2-4}
\cmidrule(lr){5-7}
&
SR $\uparrow$
&
VSF $\uparrow$
&
RT $\downarrow$
&
SR $\uparrow$
&
VSF $\uparrow$
&
RT $\downarrow$
\\
\midrule

MGT (Conf.-Refine)
& 0.69 & 0.20 & 0.03
& 0.84 & 0.58 & 0.07
\\

MGT + EmbOpt-30
& 0.75 & 0.29 & 0.08
& 0.90 & 0.74 & 0.30
\\

\midrule

Conf.-only Remask
& 0.84 & 0.49 & 0.08
& 0.92 & 0.78 & 0.43
\\

Geom.-only Remask
& 0.93 & 0.64 & 0.08
& 0.95 & 0.84 & 0.44
\\

Prob.-only Screening
& 0.88 & 0.55 & 0.07
& 0.94 & 0.81 & 0.41
\\

Independent Exact Search
& 0.91 & 0.60 & 0.08
& 0.95 & 0.82 & 0.44
\\

MGMP w/o Polish
& 0.91 & 0.64 & 0.07
& 0.94 & 0.83 & 0.40
\\

\midrule

\textbf{Full MGMP}
& \textbf{0.96} & \textbf{0.67} & \textbf{0.08}
& \textbf{0.97} & \textbf{0.87} & \textbf{0.44}
\\
\bottomrule
\end{tabular}
}

\vspace{-0.8em}
\end{wraptable}

Table~\ref{tab:ggts_ablation} isolates the main design choices in GGTS.
For \textit{where to edit}, confidence-only remasking reduces SR to
0.84/0.92 on Ring Maze/Kuka, while geometry-only remasking reaches
0.93/0.95, showing that geometric attribution provides the primary
localization signal, with confidence serving as complementary guidance. For \textit{what to fill} and \textit{how to commit},
probability-only screening and independent exact search reach
0.88/0.94 and 0.91/0.95 SR, respectively, below Full MGMP
(0.96/0.97), supporting geometry-guided screening and sequential
verification. Finally, MGT+EmbOpt-30 reaches only 0.75/0.90 SR, whereas MGMP w/o Polish already achieves 0.91/0.94. Polishing further improves this to 0.96/0.97, indicating that discrete search performs the main structural repair while continuous optimization resolves residual local infeasibilities. Additional analyses in App.~\ref{sec:ablation studyapp} examine the trade-off between first-order screening and exact verification, and sensitivity to the main refinement hyperparameters.

\vspace{-2mm}
\begin{wrapfigure}{r}{0.42\textwidth}
    \vspace{-2mm}
    \centering

    \includegraphics[width=\linewidth]{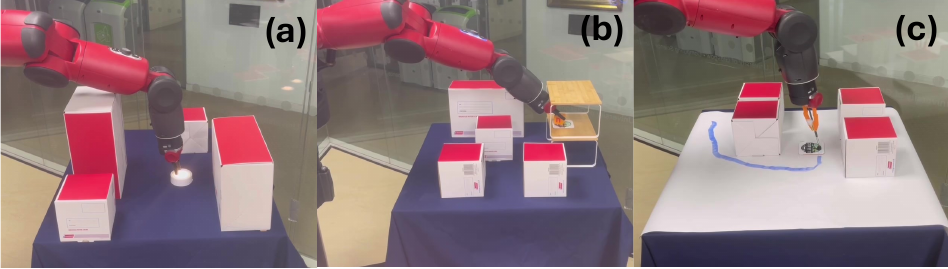}
    
    \vspace{-0.6em}

    \caption{Real-world tasks: 
    (a) button pressing; (b) shelf placing; (c) drawing.}
    \label{fig:real}

    \vspace{0.7em}

    \refstepcounter{table}
    \label{tab:real}
    {Table~\thetable: Results on real-world tasks.}

    \vspace{0.3em}
    
    \resizebox{\linewidth}{!}{%
    \begin{tabular}{@{}lcccccc@{}}
        \toprule
        \textbf{Method} &
        \multicolumn{2}{c}{\textbf{Shelf}} &
        \multicolumn{2}{c}{\textbf{Button}} &
        \multicolumn{2}{c}{\textbf{Drawing}} \\
        \cmidrule(lr){2-3}
        \cmidrule(lr){4-5}
        \cmidrule(l){6-7}
        & SR & VSF & SR & VSF & SR & VSF \\
        \midrule
        MPD & 0.533 & 0.293 & 0.733 & 0.493 & 0.733 & 0.333 \\
        PBD & 0.400 & 0.213 & 0.667 & 0.360 & 0.600 & 0.267 \\
        MGMP &
        \cellcolor{gray!15}$\mathbf{0.733}$ &
        \cellcolor{gray!15}$\mathbf{0.360}$ &
        \cellcolor{gray!15}$\mathbf{0.800}$ &
        \cellcolor{gray!15}$\mathbf{0.613}$ &
        \cellcolor{gray!15}$\mathbf{0.867}$ &
        \cellcolor{gray!15}$\mathbf{0.453}$ \\
        \bottomrule
    \end{tabular}%
    }

    \vspace{-8mm}
\end{wrapfigure}

\subsection{Real World Experiments}
\label{sec:real}
We conduct real-world tabletop experiments with a Baxter robot on three
downstream tasks: shelf placing, button pressing, and pen drawing. The planner
takes the start and goal configurations and obstacle locations as input and
outputs a joint-space trajectory. We train a task-agnostic planner from demonstrations collected with Baxter's left arm while avoiding
four tracked tabletop box obstacles. At test time, we evaluate five unseen
environments with three planning problems each, using five candidate
trajectories per method. For the 3D tasks, two training boxes are replaced
with larger unseen rectangular obstacles. Table~\ref{tab:real} shows that
MGMP performs consistently well across all tasks.
See details in App.~\ref{sec:real-world}.

\vspace{-2mm}
\section{Conclusion}
\label{sec:conclusion}


\vspace{-3mm}
We presented MGMP, which reframes generative motion planning by keeping the learned trajectory prior active beyond initial generation. A shared discrete trajectory space supports both efficient parallel generation and structural repair, while GGTS turns prior-supported motion alternatives into a geometry-guided, targeted search space. Rather than relying on local continuous refinement to discover new routes implicitly, MGMP directly restructures trajectories through discrete token replacement, using continuous optimization only for residual correction. Controlled basin-escape and route-invalidation experiments show that discrete search provides the main repair capability, while broader evaluations
demonstrate generalization across diverse planning settings. More broadly, MGMP shows how learned
trajectory priors can serve not only as generators, but also as structured proposal spaces for
geometry-guided test-time adaptation.

\subsection*{AI use statement}
We used generative AI tools solely for language editing, including improving the grammar, clarity, and phrasing of the paper. All AI-assisted edits were reviewed by the authors for technical accuracy and consistency with the intended meaning. Generative AI was not used to generate experimental results or determine the scientific conclusions of this work. The authors take full responsibility for the final content of the paper.

\subsection*{Reproducibility statement}
Experiment details, including all hyperparameters and implementation details, are in Appendix Sec.~\ref{sec:Experiment_Details_app}, with further results in Appendix Sec.~\ref{sec:Additional Simulation Results app}. These materials enable full reproducibility.

\bibliography{iclr2027_conference}
\bibliographystyle{iclr2027_conference}

\clearpage
\appendix
\appendixpage

\thispagestyle{empty}

\startcontents[app]

\medskip

\begingroup
  \setcounter{tocdepth}{3}
  \printcontents[app]{l}{1}{\setcounter{tocdepth}{3}}
\endgroup

\clearpage

\section{Additional Method Details}
\subsection{MGMP Inference with Geometry-Guided Token Search}
\label{app:algorithm}

\begin{algorithm}[H]
\setstretch{1.05}
\caption{MGMP Inference with Geometry-Guided Token Search}
\label{alg:ggts}
\begin{algorithmic}[1]

\Require Condition $\mathbf{c}=(c_s,c_g,w)$, MGT $f_\theta$,
decoder $\mathcal{D}$, codebook $\mathcal{K}$, objective $J$,  Tabu-list $\mathcal{T}$
\Require Rounds $R$, proposal size $M$, shortlist size $k$,
attribution mass $\rho$, polishing steps $P$, momentum $\mu$
\Ensure Refined trajectory $\hat{\mathbf{x}}$

\Comment{Initial masked generation}
\State $\boldsymbol{\ell}\leftarrow f_\theta([\textsc{Mask}]^N,\mathbf{c}),
\quad \mathbf{p}\leftarrow\softmax(\boldsymbol{\ell})$
\State $\mathbf{z}\leftarrow\mathrm{GumbelMax}(\boldsymbol{\ell}),
\quad c_i\leftarrow p_{i,z_i}$
\State $\mathbf{e}\leftarrow\mathcal{K}[\mathbf{z}],
\quad \hat{\mathbf{x}}\leftarrow\mathcal{D}(\mathbf{e}),
\quad \mathcal{T}\leftarrow\emptyset$

\Comment{Geometry-Guided Token Search}
\For{$r=1$ to $R$}
    \If{$\hat{\mathbf{x}}$ is feasible}
        \State \textbf{break}
    \EndIf

    \Comment{WHERE to edit}
    \State $\mathbf{e}\leftarrow\mathcal{K}[\mathbf{z}]$
    \State $s_i\leftarrow
    \|\nabla_{\mathbf{e}_i}J(\mathcal{D}(\mathbf{e}))\|_2
    \bigl(1+\lambda_{\mathrm{conf}}(1-c_i)\bigr)$
    \State $\mathcal{I}\leftarrow
    \mathrm{CumulativeTop}(s,\rho)
    \cup\mathrm{CollisionTokens}(\hat{\mathbf{x}},w)$

    \Comment{WHAT to fill}
    \State $\boldsymbol{\ell}\leftarrow
    f_\theta(\mathrm{Mask}(\mathbf{z},\mathcal{I}),\mathbf{c}),
    \quad \mathbf{p}\leftarrow\softmax(\boldsymbol{\ell})$

    \ForAll{$i\in\mathcal{I}$}
        \State $c_i\leftarrow p_{i,z_i},\quad
        \mathcal{C}_i\leftarrow\mathrm{TopM}(p_i)$
        \State $\tilde{\mathbf{p}}_i\leftarrow
        \softmax\!\left(
        \frac{\boldsymbol{\ell}_i[\mathcal{C}_i]+\mathbf{g}_i}{\tau}
        \right),
        \quad \mathbf{g}_i\sim\mathrm{Gumbel}(0,1)$
        \State $\bar{\mathbf{e}}_i\leftarrow
        \sum_{j\in\mathcal{C}_i}\tilde p_{i,j}\mathcal{K}[j]$
        \State $\mathcal{S}_i\leftarrow
        \mathrm{TopK}_{j\in\mathcal{C}_i}
        \left[
        -\tilde p_{i,j}
        \nabla_{\bar{\mathbf{e}}_i}J
        \cdot(\mathcal{K}[j]-\bar{\mathbf{e}}_i)
        \right]$
    \EndFor

    \Comment{Exact sequential verification}
    \ForAll{$i\in\mathrm{SortDesc}(\mathcal{I},s_i)$}
        \State $j^\star\leftarrow
        \arg\min_{\substack{j\in\mathcal{S}_i\\(i,j)\notin\mathcal{T}}}
        J\!\left(\mathcal{D}(\mathcal{K}[\mathbf{z}^{(i\leftarrow j)}])\right)$
        \If{$J(\mathcal{D}(\mathcal{K}[\mathbf{z}^{(i\leftarrow j^\star)}]))
        <J(\hat{\mathbf{x}})$}
            \State $z_i\leftarrow j^\star,\quad
            \hat{\mathbf{x}}\leftarrow\mathcal{D}(\mathcal{K}[\mathbf{z}]),
            \quad \mathcal{T}\leftarrow\emptyset$
        \Else
            \State $\mathcal{T}\leftarrow
            \mathcal{T}\cup\{(i,j):j\in\mathcal{S}_i\}$
        \EndIf
    \EndFor
\EndFor

\Comment{Optional continuous polishing}
\State $\mathbf{e}\leftarrow\mathcal{K}[\mathbf{z}],
\quad \mathbf{m}\leftarrow\mathbf{0}$
\For{$t=1$ to $P$}
    \State $\mathbf{m}\leftarrow
    \mu\mathbf{m}+\nabla_{\mathbf{e}}J(\mathcal{D}(\mathbf{e}))$
    \State $\mathbf{e}\leftarrow\mathbf{e}-\mathbf{m}$
\EndFor
\State \Return $\mathcal{D}(\mathbf{e})$

\end{algorithmic}
\end{algorithm}

\subsection{Planning Objectives for Different Task Settings}
\label{sec:task_scenarios}
After training, our model can handle several general planning scenarios. 

\paragraph{Single-agent planning.} In this scenario, the configuration $x$ is from a single point or a single robotic arm. Generative planning only needs to consider the start/end distance error and scene collisions over the whole length $T$ of the trajectory. The objective function is defined as follows:
\begin{equation}
    \mathcal{L} =
    \sum_{t=1}^T \lambda_{c}\mathcal{L}_{\mathcal{SDF}}(\hat{x}_{t}, w) + \lambda_{sg}\mathcal{L}_{sg}(\hat{x}, c_s, c_g)
    \label{eq:sinopt}
\end{equation}
where $\lambda_{*}$ are weight factors for the start/goal error $\mathcal{L}_{sg}$ and collision error $\mathcal{L}_{\mathcal{SDF}}$.
The former is an L1 loss on configurations, while the latter is given by
$
    \mathcal{L}_{\mathcal{SDF}}=\mathrm{ReLU}(-\mathrm{SDF}(\hat{x},w)+\epsilon)
    \label{eq:sdf}
$
where $\epsilon$ defines a margin around objects.

\paragraph{Dual-agent planning.} For dual-agent or dual-arm planning tasks, in addition to avoiding collisions with the environment, it is also necessary to prevent collisions between the two agents/arms. Accordingly, a self-collision term is introduced to the optimization objective:
\begin{equation}
    \mathcal{L} =
    \sum_{t=1}^{T}(\lambda_{c}\mathcal{L}_\mathcal{SDF}(\hat{x}_t^0, w)+\lambda_{c}\mathcal{L}_\mathcal{SDF}(\hat{x}_{t}^1, w)+\lambda_{s} \mathcal{L}_\mathcal{SDF}(\hat{x}_t^0,\hat{x}_{t}^1))+\lambda_{sg}\mathcal{L}_{sg}
    \label{eq:dualopt}
\end{equation}
where $\lambda_{s}$ is the weight for self-collision error, $\hat{x}_t^0$ and  $\hat{x}_t^1$ are the trajectory of each agent at frame $t$. 

\paragraph{Planning with additional obstacles.}
Our model can perform motion planning with a larger number of obstacles than seen in its training data. We partition the set of obstacles $\left \{w\right \}$ into $J$ subsets, where the cardinality of each does not exceed the maximum number seen by the obstacle encoder during training. The problem is then reformulated as solving over the joint solution space conditioned on separate obstacle sets. The model conditioned on each subset can be viewed as an expert over those obstacles, and the trajectory constraints for the full set are given by multiplying these, similar to a product of experts: 
$
e=\sum_j^Je_{j}$,
$
p(z \mid \left \{w\right \}) \propto \prod_{j=1}^{J}  p(z \mid w_j)
$
Based on this, we predict logits for each set individually and sum them to obtain the final global-condition logits. These logits are then sent into the softmax operator to calculate the final probability distribution and the codes. The optimization objective for the codes is the same as in the single-agent task.

\section{Experiment Details}
\label{sec:Experiment_Details_app}

\subsection{Hyperparameters}
\label{sec:hyperparameters}

We report the hyperparameters used for MGMP on Maze2D, 7-DoF Kuka, and 14-DoF DualKuka. Ring Maze uses the same configuration as Maze2D, while Controlled Route Invalidation uses the same configuration as the 7-DoF Kuka experiments. 
No additional hyperparameter tuning is performed on Ring Maze or Controlled Route Invalidation.
MGMP consists of a VQ-VAE trajectory tokenizer, a masked generative transformer, and test-time token-space refinement followed by short continuous polishing.
Tables~\ref{tab:vq_hyperparams}, \ref{tab:mgt_hyperparams}, and \ref{tab:refine_hyperparams} summarize the corresponding hyperparameters.

\raggedbottom

\begin{table}[H]
\centering

\caption{VQ-VAE hyperparameters.}
\label{tab:vq_hyperparams}

\begin{tabular}{@{}lccc@{}}
\toprule
\textbf{Parameter} &
\textbf{Maze2D} &
\textbf{7-DoF Kuka} &
\textbf{14-DoF DualKuka} \\
\midrule
Codebook size & 8192 & 8192 & 8192 \\
Code dimension & 16 & 256 & 256 \\
Encoder width & 16 & 256 & 256 \\
Encoder depth & 4 & 4 & 4 \\
Downsampling rate & 2 & 2 & 1 \\
Stride size & 2 & 2 & 2 \\
Dilation growth rate & 3 & 3 & 3 \\
Activation & SiLU & SiLU & SiLU \\
Quantizer & ema\_reset & ema\_reset & ema\_reset \\
EMA decay $\mu$ & 0.95 & 0.95 & 0.95 \\
Reconstruction loss & smooth-$\ell_1$ & smooth-$\ell_1$ & smooth-$\ell_1$ \\
Commit weight & 0.02 & 0.02 & 0.02 \\
Velocity loss weight & 0.5 & 0.5 & 0.5 \\
FK loss weight & - & 0.6 & 0.6 \\
Learning rate & $3{\times}10^{-4}$ & $3{\times}10^{-4}$ & $1{\times}10^{-4}$ \\
LR milestones & [250k, 400k, 600k] & [250k, 400k, 600k] & [250k, 400k, 600k] \\
LR decay $\gamma$ & 0.1 & 0.1 & 0.1 \\
Warmup iterations & 1000 & 1000 & 1000 \\
Weight decay & 0.0 & 0.0 & $1{\times}10^{-6}$ \\
Training iterations & 800k & 800k & 800k \\
Batch size & 256 & 256 & 256 \\
\bottomrule
\end{tabular}

\vspace{1.2em}

\caption{Masked generative transformer hyperparameters.}
\label{tab:mgt_hyperparams}

\begin{tabular}{@{}lccc@{}}
\toprule
\textbf{Parameter} &
\textbf{Maze2D} &
\textbf{7-DoF Kuka} &
\textbf{14-DoF DualKuka} \\
\midrule
Embedding dimension & 256 & 512 & 512 \\
Number of layers & 8 & 8 & 8 \\
Number of heads & 16 & 16 & 16 \\
FFN multiplier & 4 & 4 & 4 \\
Dropout & 0.2 & 0.2 & 0.2 \\
Block size & 16 & 16 & 32 \\
Learning rate & $1{\times}10^{-4}$ & $1{\times}10^{-4}$ & $1{\times}10^{-4}$ \\
Weight decay & $1{\times}10^{-6}$ & $1{\times}10^{-4}$ & $1{\times}10^{-4}$ \\
Optimizer & AdamW & AdamW & AdamW \\
Training iterations & 200k & 200k & 200k \\
Batch size & 256 & 256 & 256 \\
\bottomrule
\end{tabular}

\end{table}

\begin{table}[H]
\centering

\caption{
Test-time GGTS and continuous polishing hyperparameters.
During verification, the number of remasked tokens is capped at
40\% of the trajectory tokens; for the Kuka setting, this corresponds
to at most four tokens. Continuous polishing uses momentum-accumulated
raw gradients with no separate learning rate. The SDF and SG weights
are switched to their post-convergence values once the SDF loss falls
below $5{\times}10^{-4}$.
}
\label{tab:refine_hyperparams}

\begin{tabular}{@{}lccc@{}}
\toprule
\textbf{Parameter} &
\textbf{Maze2D} &
\textbf{7-DoF Kuka} &
\textbf{14-DoF DualKuka} \\
\midrule
GGTS refinement rounds
    & 2 & 2 & 2 \\

Token updates per round
    & 2 & 2 & 2 \\

Proposed code candidates per update
    & 64 & 16 & 16 \\

Candidates retained after screening
    & 32 & 2 & 2 \\

Maximum verification remask ratio
    & 0.4 & 0.4 & 0.4 \\

Polishing iterations
    & 5 & 5 & 5 \\

Momentum
    & 0.9 & 0.9 & 0.9 \\

SDF weight, pre-convergence
    & 200 & 300 & 300 \\

SG weight, pre-convergence
    & 100 & 150 & 150 \\

SDF weight, post-convergence
    & 400 & 600 & 600 \\

SG weight, post-convergence
    & 100 & 450 & 450 \\
\bottomrule
\end{tabular}

\end{table}

For the standard Maze2D, Kuka, and DualKuka environments, we randomly split the dataset into training and validation sets with a 9:1 ratio. 
All hyperparameters are selected by grid search on the corresponding held-out validation set for each benchmark.
For the VQ-VAE trajectory tokenizer, we choose the configuration that achieves the lowest validation reconstruction error. 
For the masked generative transformer and test-time refinement, we select the configurations that achieve the highest validation success rate.

\subsection{Network Architecture}
\label{app:network_architecture}

MGMP consists of two trainable network components: a VQ-VAE trajectory tokenizer and a masked generative transformer. The tokenizer learns a compact discrete representation of continuous trajectories, while the transformer models the conditional distribution over trajectory tokens given the start configuration, goal configuration, and obstacle representation.

\paragraph{Trajectory tokenizer.}
The trajectory tokenizer is implemented as a 1D-CNN VQ-VAE with an encoder–decoder architecture. The encoder is composed of 3 temporal convolutional blocks, each consisting of a 1D-CNN layer and a residual 1D-CNN layer. The kernel size of the 1D-CNN layer is 4.
The decoder shares the mirrored structure with the encoder. The EMA tokenizer is selected as the tokenizer for the trajectory codes. The temporal downsampling factor for the single-agent task is $N/T$ = $1/4$, which means four trajectory steps will be encoded into 1 token. For the dual-agent task, this ratio is set as $1/2$ for better reconstruction results. Objective is:

\[
\mathcal{L}
=
\lambda_{\mathrm{rec}}\mathcal{L}_{\mathrm{smoothL1}}
+
\beta\|\hat{\mathbf e}-\mathrm{sg}[\mathbf e]\|_2^2
+
\lambda_v\mathcal{L}_{\mathrm{vel}}
+
\lambda_{\mathrm{FK}}\mathcal{L}_{\mathrm{FK}}.
\]
The forward-kinematics loss is applied only to the Kuka and DualKuka
trajectories; for Maze2D, $\lambda_{\mathrm{FK}}=0$.

\paragraph{Masked generative transformer.}
The masked generative transformer consists of two components. The first is a lightweight obstacle encoder that takes the obstacle center coordinates $w_C \in \mathbb{R}^{n\times 3}$ as input. This encoder is built on 3-layer self-attention blocks without positional embedding. The hidden dimension and the output dimension are 256. 

The masked generative transformer (MGT) backbone consists of 8 self-attention-based Transformer blocks. The MGT hidden dimension is 256 for Maze2D and 512 for Kuka/DualKuka. The start/goal condition is projected by a 2-layer MLP. The obstacle embedding is injected into the transformer through an AdaLN block. We apply conditional modulation only to the first six Transformer layers, while the final two layers serve as the output head.

\subsection{Simulation Dataset and Training Data}
\label{app:env_details}

We use BIT*~\citep{gammell2015batch} to generate expert trajectories for all simulated environments using Pybullet simulation~\citep{coumans2016pybullet}. Each training sample is a feasible solution connecting a randomly sampled start-goal pair under a fixed obstacle layout. For each task, we construct 500 training environments and generate 400 trajectories per environment, resulting in 200K trajectory samples per task. We developed our task settings from ~\citep{luopotential}. All training trajectories are collision-free BIT*~\citep{gammell2015batch} solutions. Evaluation environments and start-goal pairs are disjoint from the training set.

\paragraph{Ring Maze}
Each Ring Maze environment is a $5\times5$\,m 2-D workspace containing seven near-enclosing C-shaped obstacles, each with a single opening oriented along one of the four cardinal directions. This construction creates enclosed regions with narrow exits that may require substantial route changes, while the obstacle configuration remains fixed throughout each planning problem. We generate 500 training, 20 validation, and 100 unseen test environments. Each training environment contains 400 BIT*-generated expert planning problems, yielding 200K training trajectories. We sample 10 problems per validation environment and 40 per test environment, resulting in 200 validation and 4,000 test problems. Because uniform start--goal sampling rarely places endpoints inside the ring cavities, we use a trap-aware sampler covering \emph{ring-to-ring}, \emph{ring-to-free}, \emph{free-to-ring}, and \emph{free-to-free} queries, where ``ring'' denotes a collision-free state inside a C-shaped enclosure and ``free'' denotes a state in the remaining free workspace. Under this protocol, $70\%$ of the test problems require escaping at least one enclosure. BIT* is used to generate the training trajectories; start and goal states are checked for collision validity, and only problems for which BIT* finds a feasible solution are retained.

\paragraph{Maze2D.}
Maze2D is a low-dimensional planning task in which a holonomic point robot moves in a two-dimensional workspace. The robot state is its planar position, and the planner must generate a collision-free path from a start state to a goal state while avoiding square or concave obstacles. Collision is determined by whether the point robot lies inside an obstacle region. The dataset contains feasible paths generated by BIT*~\citep{gammell2015batch} under randomly sampled obstacle layouts and start-goal pairs.

\paragraph{Kuka.}
Kuka evaluates joint-space motion planning for a 7-DoF manipulator in cluttered 3D workspaces. Each training environment contains randomly placed box obstacles around the robot. For each environment, BIT*~\citep{gammell2015batch} generates feasible joint-space trajectories between randomly sampled start and goal configurations. The resulting trajectories are used to train the trajectory tokenizer and masked generative transformer.

\paragraph{DualKuka.}
DualKuka extends the Kuka setting to two coupled 7-DoF manipulators operating in a shared workspace. The state dimension is therefore 14, and feasible trajectories must avoid both workspace obstacles and inter-arm collisions. As in the Kuka setting, each environment contains randomly placed obstacles, and BIT*~\citep{gammell2015batch} is used to generate feasible trajectories for randomly sampled start-goal pairs.

\paragraph{Trajectory-level training samples.}
We preserve trajectory boundaries when constructing training samples. Unlike stride-one sliding-window augmentation over concatenated trajectories, our training samples are extracted from individual BIT*~\citep{gammell2015batch} solutions. This avoids windows that cross different planning problems and therefore do not correspond to a valid trajectory for a single start-goal pair. Each training sequence is thus consistent with its own start configuration, goal configuration, and obstacle layout.

\subsection{Baseline Details}
\label{app:baseline_details}

We compare MGMP with eleven representative baselines, covering both classical and learning-based planning methods.

\paragraph{Classical planners.}
RRT*~\citep{karaman2011sampling}, P-RRT*~\citep{qureshi2016potential}, and BIT*~\citep{gammell2015batch} are sampling-based planners with different exploration strategies; BIT* additionally combines informed sampling with graph search to progressively improve candidate paths.
RMP~\citep{ratliff2018riemannian} is a reactive motion policy that combines goal-reaching and obstacle-avoidance objectives online.
We further include three trajectory-optimization baselines: CHOMP~\citep{ratliff2009chomp}, which performs gradient-based optimization of collision and smoothness objectives; TrajOpt~\citep{schulman2014motion}, which optimizes trajectories through sequential convex optimization; and STOMP~\citep{kalakrishnan2011stomp}, which uses stochastic trajectory perturbations to minimize planning costs. All classical planners are given a maximum planning budget of 5,s per problem.

\paragraph{Learning-based planners.}
M$\pi$Nets~\citep{fishman2023motion} learns a neural motion policy conditioned on the planning goal and environment.
Motion Planning Diffusion (MPD)~\citep{carvalho2023motion} models a trajectory prior with diffusion and incorporates planning costs during inference to guide sampled trajectories.
Potential-Based Diffusion (PBD)~\citep{luopotential} combines diffusion generation with potential-based guidance for goal reaching and obstacle avoidance.
SafeFlowMatcher~\citep{yang2026safeflowmatcher} uses flow matching for trajectory generation and a prediction--correction procedure with control barrier functions to continuously correct generated paths toward the safe set.

\subsection{Running time}
\label{app:plan_time}
All experiments are run on a single workstation with an AMD Ryzen 9
5950X CPU and one NVIDIA RTX 4090 GPU. Device assignment follows the
implementation of each planner: sampling-based planners (RRT*, P-RRT*,
and BIT*) use CPU implementations based on OMPL, while learning-based
planners and trajectory-optimization baselines use GPU. Candidate-based methods are evaluated with 20
trajectories per planning problem. Regardless of the internal collision
objective, guidance mechanism, or optimizer used by a method, all
returned trajectories are certified using the same final collision
checker and endpoint-validity criteria. Sampling-based planners return
a single solution and therefore do not report valid-sample fraction.

Planning time (PT) measures the wall-clock time from planner invocation
until the first trajectory is certified as feasible, or until the
maximum planning budget is reached. It includes test-time sampling,
optimization or guidance, collision-distance and forward-kinematics
evaluations where applicable, and the validity checks required to
certify the returned solution. Running time (RT), in contrast, measures
the wall-clock cost of generating and refining the complete set of 20
candidate trajectories. One-time model and dataset loading and the
post-hoc validation used solely to compute valid-sample fraction are
excluded from both timing measurements. Results are reported as
mean $\pm$ standard deviation across test environments, after
aggregating the planning queries within each environment.

\subsection{Experiment Details for Ring Maze}
\label{sec:ring_impl}

For Ring Maze, all candidate-based methods are evaluated using 20 trajectories per planning problem under the common validity criteria to compare with learning-based motion planners. For the trajectory-optimization baselines CHOMP~\citep{ratliff2009chomp}, TrajOpt~\citep{schulman2014motion}, and STOMP~\citep{kalakrishnan2011stomp}, we additionally control the trajectory initialization. All three methods receive the same set of 20 initial trajectories for each query: one straight-line configuration-space interpolation between the start and goal and 19 randomized cubic B\'ezier trajectories with the same fixed endpoints. The two interior B\'ezier control points are sampled within the workspace bounds using deterministic per-query random seeds and do not depend on the obstacle configuration. The initial trajectories use the same waypoint discretization. CHOMP, TrajOpt, and STOMP then independently refine this common candidate set using their respective optimization procedures. For STOMP, the internal stochastic rollouts used to estimate an update are not counted as additional planning candidates. Method-specific hyperparameters for the trajectory optimizers are selected on held-out validation environments that are disjoint from the test environments, under the same 5\,s maximum planning budget used at test time, and are fixed during evaluation. MPD~\citep{carvalho2023motion}, PBD~\citep{luopotential}, and SafeFlowMatcher~\citep{yang2026safeflowmatcher} each generate 20 candidates using their native sampling procedures and apply their respective test-time refinement mechanisms. For the MGMP variants, MGT+EmbOpt-30 uses the same MGT-generated candidates as MGMP but replaces discrete token refinement with 30 steps of continuous embedding-space optimization. MGMP w/o Polish applies GGTS without the final continuous polishing stage, while full MGMP applies GGTS followed by the short continuous polishing stage.

\subsection{Implementation Details for Controlled Route Invalidation on Kuka}
\label{app:control_kuka}

For each controlled route-invalidation problem, we first obtain a verified
feasible trajectory $x_A$ (Route A). To remove representation mismatch between
continuous- and token-based refinement methods, we encode $x_A$ with the MGMP
tokenizer and decode the resulting hard-token sequence $z_A$ to obtain a common
reconstructed trajectory
\[
\tilde{x}_A = D(E(x_A)).
\]
We retain only cases for which $\tilde{x}_A$ remains feasible in the original
scene. We then relocate one obstacle to invalidate $\tilde{x}_A$ while
preserving an alternative feasible route.

All methods receive the same invalidated trajectory $\tilde{x}_A$ and the same
modified obstacle configuration. CHOMP, TrajOpt, STOMP, and
SafeFlowMatcher-Corr. are initialized directly from $\tilde{x}_A$, with the
prediction stage of SafeFlowMatcher bypassed.

For PBD, we use its native trajectory-refinement procedure rather than fresh
generation: $\tilde{x}_A$ is forward-noised to diffusion step
$k_{\mathrm{PBD}}=3$ and denoised under the modified constraints, after which
collision-free sections of the denoised trajectory replace the corresponding
invalid sections of the current trajectory. This process is repeated for at most
$R_{\mathrm{PBD}}=5$ refinement attempts, following the original Kuka
evaluation setting.

For MPD, whose standard inference starts from Gaussian noise, we use an adapted
warm-start variant. We forward-noise $\tilde{x}_A$ to diffusion step
$k_{\mathrm{MPD}}=10$ using the original MPD diffusion process and then apply its
standard cost-guided reverse denoising under the modified obstacle configuration.
Start and goal configurations are clamped throughout denoising. MPD-WarmStart
therefore uses the same invalidated initial trajectory as the other methods
rather than generating a new trajectory from pure Gaussian noise.

The diffusion noise levels and refinement budgets are selected on held-out
validation environments and fixed for all test problems.
MGT + Continuous EmbOpt is initialized from the token embeddings corresponding
to $z_A$ and applies 30 steps of continuous embedding-space optimization.
MGMP and MGMP w/o Polish start from the same hard-token sequence $z_A$ and
apply GGTS; the conditional MGT is used only to propose replacement tokens
during refinement, rather than to regenerate the initial trajectory. Full MGMP
additionally applies the final continuous polishing stage.

Thus, every method is conditioned on the same invalidated trajectory under
identical modified constraints, and no method is allowed to generate a new
trajectory from scratch. This construction isolates trajectory-repair capability
from differences in initial-generation quality. The average per-step
$\ell_2$ reconstruction error between $x_A$ and $\tilde{x}_A$ is approximately
$1\times10^{-4}$, and reconstruction does not alter the feasibility of any
retained Route A before obstacle relocation, making tokenizer-induced differences
negligible in this comparison.

For each successful repair, we compare the repaired trajectory with the shared invalidated initial trajectory in joint space. At each timestep, we compute the $\ell_2$ distance between the corresponding joint configurations. Average deviation is the mean of these distances over the trajectory, while maximum deviation is the largest distance at any timestep. A waypoint is considered changed when its joint-space deviation exceeds $\epsilon=0.1$ rad. Changed span is the longest contiguous sequence of changed waypoints, reported as a percentage of the full trajectory length. Larger deviation values indicate greater changes in joint configuration, while a larger changed span indicates that the repair modifies a broader portion of the route rather than only the local collision region. 

\subsection{Implementation Details for Unseen Obstacle Layouts}
\label{app:impl_unseen_layouts}

\paragraph{Maze2D.}
\label{app:impl_maze2d}

For Maze2D, the VQ-VAE tokenizer uses a 1D ResNet encoder with channel width 16 and depth 4. It applies a temporal downsampling factor of 4 and uses a codebook of 8192 entries with code dimension 16. The tokenizer is trained with a smooth-$\ell_1$ reconstruction loss, EMA-reset vector quantization, commitment weight 0.02, and an additional velocity reconstruction loss with weight 0.5. We train the VQ-VAE for 800k iterations with batch size 256 using AdamW and an initial learning rate of $3{\times}10^{-4}$, decayed by $0.1$ at 250k, 400k, and 600k iterations. Since the trajectory horizon is $H=48$, each trajectory is compressed into 12 discrete tokens.

The masked generative transformer contains 8 self-attention layers, embedding dimension 256, 16 attention heads, FFN multiplier 4, and dropout 0.2. The six rectangular obstacles are represented by their 2D center coordinates and encoded using a ViT-1D obstacle encoder. During training, we apply obstacle-conditioning dropout with probability 0.2. The transformer is trained for 200k iterations with batch size 256 using AdamW, learning rate $1{\times}10^{-4}$, and weight decay $1{\times}10^{-6}$.

At inference time, the refinement objective is
$J=w_{\mathrm{SDF}}L_{\mathrm{SDF}}+w_{\mathrm{sg}}L_{\mathrm{sg}}$.
Before SDF convergence, defined as $\mathcal{L}_{\mathrm{sdf}}<5{\times}10^{-4}$, we use $(w_{\mathrm{sdf}},w_{\mathrm{sg}})=(200,100)$; after convergence, we switch to $(400,100)$.

\paragraph{7-DoF Kuka and 14-DoF DualKuka.}
For the Kuka environments, the VQ-VAE uses a 1D ResNet encoder with width 256 and depth 4, together with a codebook of 8192 entries and code dimension 256. The 7-DoF Kuka uses a temporal downsampling factor of 4, producing 12 tokens per trajectory. The 14-DoF DualKuka uses a smaller downsampling factor of 2, producing 24 tokens per trajectory to preserve finer temporal resolution for the higher-dimensional system. In addition to the smooth-$\ell_1$ reconstruction and velocity losses, we include a forward-kinematics Cartesian position loss with weight 0.6, computed every 10 steps, to improve the spatial accuracy of reconstructed joint-space trajectories. The VQ-VAE is trained for 800k iterations with batch size 256 using AdamW and an initial learning rate of $3{\times}10^{-4}$ ($1{\times}10^{-4}$ for Dualkuka), decayed by $0.1$ at 250k, 400k, and 600k iterations, with no weight decay (weight decay for Dualkuka is $1{\times}10^{-6}$).

The masked generative transformer uses 8 self-attention layers, embedding dimension 512, 16 attention heads, FFN multiplier 4, and dropout 0.2. For 7-DoF Kuka, four box obstacles are represented by their 3D center coordinates and encoded using a ViT-1D obstacle encoder. For 14-DoF DualKuka, five box obstacles are encoded in the same way. During training, obstacle conditioning is randomly dropped with probability 0.2. The transformer is trained for 200k iterations with batch size 256 using AdamW, learning rate $1{\times}10^{-4}$ and weight decay $1{\times}10^{-4}$.

At inference time, MGMP performs two token refinement steps, followed by 5 final latent-embedding optimization steps using momentum $\beta=0.9$. The SDF-guided refinement weights are set to $(w_{\mathrm{sdf}}, w_{\mathrm{sg}})=(300,150)$ before convergence and switched to $(600,450)$ after SDF convergence.

\subsection{Generalization to Additional-Obstacle Environments Setup}\label{sec:Additional-Obstacle Environments Set up}

\paragraph{Construction of Additional-Obstacle Environments.}
We construct additional-obstacle evaluation environments by augmenting held-out base scenes rather than sampling new layouts from scratch. The original obstacles are inherited from the corresponding base evaluation environment, and additional obstacles are inserted to increase the constraint density. Additional obstacles are placed using rejection sampling under non-overlap constraints with the inherited obstacles and invalid workspace regions. This avoids degenerate or physically invalid layouts that could arise from independently sampling all obstacle components, such as overlapping obstacles or newly added obstacles occluding predefined free-space regions. The resulting environments are used only for evaluation; no training trajectories are collected for the combined obstacle configurations. Each start-goal problem is pre-verified with BIT* under a 60-second time budget, ensuring that all reported failures correspond to planner limitations rather than infeasible planning instances.

\subsection{Generalization to Unseen Obstacle Geometries Environments Setup}
\label{sec:Unseen Obstacle Geometries Environments Set up}

\begin{figure*}
   \centering
    \includegraphics[width=0.9\linewidth]{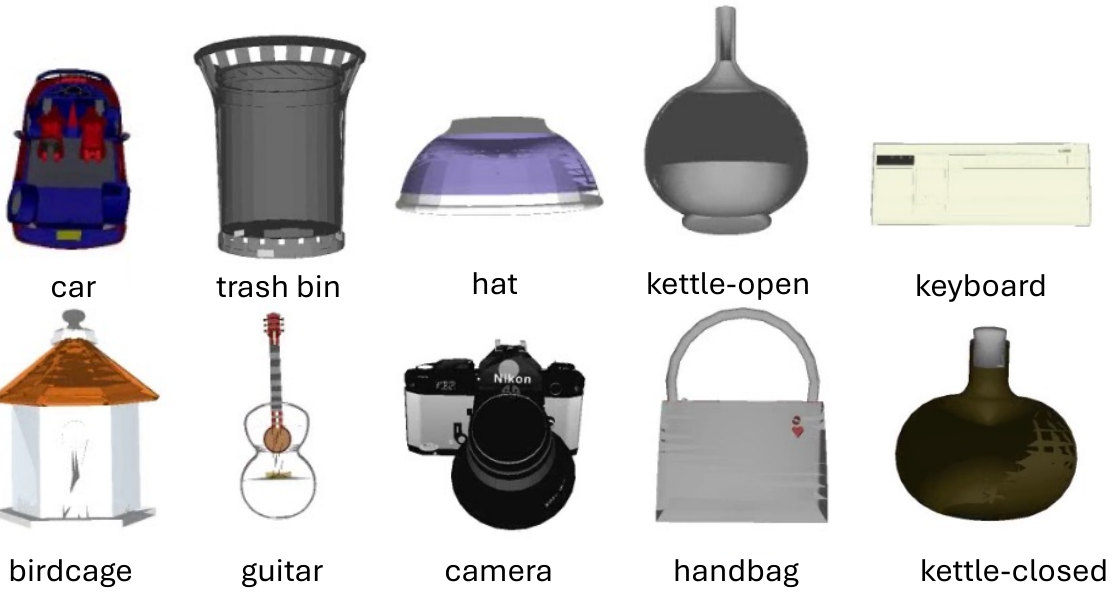}
   \caption{Object-level mesh obstacles used for evaluating generalization to unseen obstacle geometries. These objects replace the box obstacles used during training, creating test environments with irregular and previously unseen obstacle shapes.}
   \label{fig:unseen_geo_objects}
\end{figure*}

To evaluate generalization to unseen obstacle geometries, we replace the box obstacles used during training with object-level meshes at inference time. The mesh library contains ten ShapeNet objects: car, trash bin, hat, open kettle, keyboard, birdcage, guitar, camera, handbag, and closed kettle, as shown in Fig.~\ref{fig:unseen_geo_objects}. These objects introduce irregular obstacle boundaries and non-cuboid geometries that are not observed during training. This setting therefore tests whether the planner can generalize to more realistic object-shaped obstacles without additional training.

Each object is represented by a triangulated mesh normalized to fit within a unit cube and then scaled in simulation using an object-specific scalar. The scale is chosen such that the resulting obstacle is larger than the box obstacles used during training. Specifically, the target world-space half-extent is set to approximately $0.34$m, corresponding to $1.7\times$ the training box half-extent of $0.20$m. Thus, this setting evaluates not only shape generalization, but also robustness to larger obstacle geometries than those observed during training.

The environment and problem generation protocol follows the additional-obstacle setting. We use the same number of environments and start-goal problems, and sample obstacle center positions from the same workspace placement rules used for the additional-obstacle experiments. The orientation of each mesh object is fixed across environments, while only its center position varies. This ensures that the main distribution shift comes from obstacle geometry and size rather than from a different placement strategy. For 7-DoF Kuka, the four box obstacle slots are replaced by four mesh objects. For 14-DoF DualKuka, we additionally include the fifth mesh obstacle. All start and goal configurations are generated using the same feasibility procedure as in the additional-obstacle experiments.

For SDF-guided refinement, we use DeepSDF to provide a differentiable signed-distance representation for each mesh object. Each object is associated with a fixed latent code $z \in \mathbb{R}^{64}$ from a precomputed latent-code library. The library contains ten latent codes for the full ShapeNet object set, from which four codes are selected per Kuka scene and five per DualKuka scene. The latent codes are fixed during planning; we do not optimize the object latent code at test time.

Given a query point $x_w$ in world coordinates and an obstacle centered at $c$, we first transform the point into the canonical coordinate frame used by DeepSDF:
\begin{equation}
    x_{\mathrm{obj}}
    =
    R\left(\frac{x_w - c}{s} - m\right) + t,
\end{equation}
where $s$ is the object scale, $m \in \mathbb{R}^{3}$ is the AABB centroid of the normalized mesh, and $(R,t)$ denotes the fixed rigid alignment between the normalized mesh frame and the DeepSDF training frame. The AABB-centroid correction accounts for meshes whose normalized geometry is not centered at the origin. The world-space signed distance is then recovered by scaling the canonical DeepSDF output:
\begin{equation}
    d(x_w) = s \cdot f_{\theta}(z, x_{\mathrm{obj}}),
\end{equation}
where $f_{\theta}$ is the DeepSDF decoder. Negative values indicate that the query point lies inside the object, while positive values indicate free space.

During planning, DeepSDF is used to compute differentiable collision losses for SDF-guided code refinement. Final trajectory validity is evaluated separately in PyBullet. This separation allows the planner to use smooth SDF gradients for refinement while retaining a physics-engine-based collision check for evaluation.

\section{Additional Simulation Results}\label{sec:Additional Simulation Results app}

This section provides qualitative results for the simulated planning tasks. Each figure compares representative trajectories generated by BIT*, PBD, and MGMP. These visualizations complement the quantitative results in the main paper by showing the geometric quality of the planned trajectories.

\subsection{Further Analysis on Ring Maze}
\label{app:ring_analysis}

We provide additional analyses to better understand when local refinement
succeeds or fails in the Ring Maze benchmark.

\paragraph{Effect of refinement budget.}
Table~\ref{tab:ring_budget} studies whether the performance gap can be closed simply by increasing the amount of local refinement. For CHOMP, increasing the number of optimization iterations from 10 to 50 has almost no effect: the success rate remains at $73.3\%$, while the valid-sample fraction varies only between $22.3\%$ and $23.7\%$. Increasing the refinement budget therefore does not substantially change the solutions reached by the optimizer. For learned guidance, most of the improvement occurs within the first three refinement steps. PBD improves substantially during the first three steps, but the gains become small thereafter: from step 3 to step 5, success increases only from $63.05\%$ to $67.37\%$. In contrast, MPD peaks after three guidance steps ($14.40\%$ success) and degrades with further refinement. These results indicate that simply increasing the continuous-refinement budget does not reliably overcome the planning basins encountered in Ring Maze.

\paragraph{Effect of collision severity.}
We further compare PBD and MGMP as the initial trajectory moves farther from feasibility. We measure collision severity using the minimum clearance of the initial trajectory to the obstacles, where negative values indicate penetration. We group trajectories into three ranges: $[-0.05, 0)$, $[-0.15, -0.05)$, and $[-0.30, -0.15)$. PBD repairs $46.5\%$, $9.6\%$, and $5.2\%$ of trajectories in these three ranges, respectively, whereas MGMP achieves $93.8\%$, $84.1\%$,
and $71.4\%$. The widening gap with deeper collisions suggests that gradient-based local refinement is effective near the feasible region, whereas GGTS remains more effective when larger trajectory modifications are required.

\begin{table}[t]
\centering
\caption{
Effect of refinement budget on Ring Maze.
SR and VSF denote success rate and valid-sample fraction, respectively.
}
\label{tab:ring_budget}
\small
\setlength{\tabcolsep}{8pt}

\begin{tabular}{lcccc}
\multicolumn{5}{c}{\textbf{(a) CHOMP: optimization iterations}} \\
\midrule
Iterations & 10 & 20 & 30 & 50 \\
\midrule
SR (\%) $\uparrow$ & 73.30 & 73.30 & 73.30 & 73.30 \\
VSF (\%) $\uparrow$ & 22.30 & 23.30 & 23.70 & 23.30 \\
\bottomrule
\end{tabular}

\vspace{0.8em}

\begin{tabular}{lcccccc}
\multicolumn{7}{c}{\textbf{(b) Learned refinement: number of guidance steps}} \\
\midrule
Steps & 0 & 1 & 2 & 3 & 4 & 5 \\
\midrule
PBD SR (\%) $\uparrow$
& 40.95 & 53.63 & 59.30 & 63.05 & 65.55 & 67.37 \\
PBD VSF (\%) $\uparrow$
& 7.80 & 12.67 & 15.82 & 18.14 & 21.95 & 26.00 \\
\midrule
MPD SR (\%) $\uparrow$
& 2.00 & 7.80 & 11.60 & \textbf{14.40} & 10.30 & 9.30 \\
MPD VSF (\%) $\uparrow$
& 0.27 & 1.00 & 1.62 & \textbf{2.25} & 1.85 & 1.80 \\
\bottomrule
\end{tabular}
\end{table}

\subsection{Generalization to Unseen Obstacle Layouts}
\label{app:per_env_unseen_layouts}

\begin{figure*}[t]
   \centering
    \includegraphics[width=1.0\linewidth]{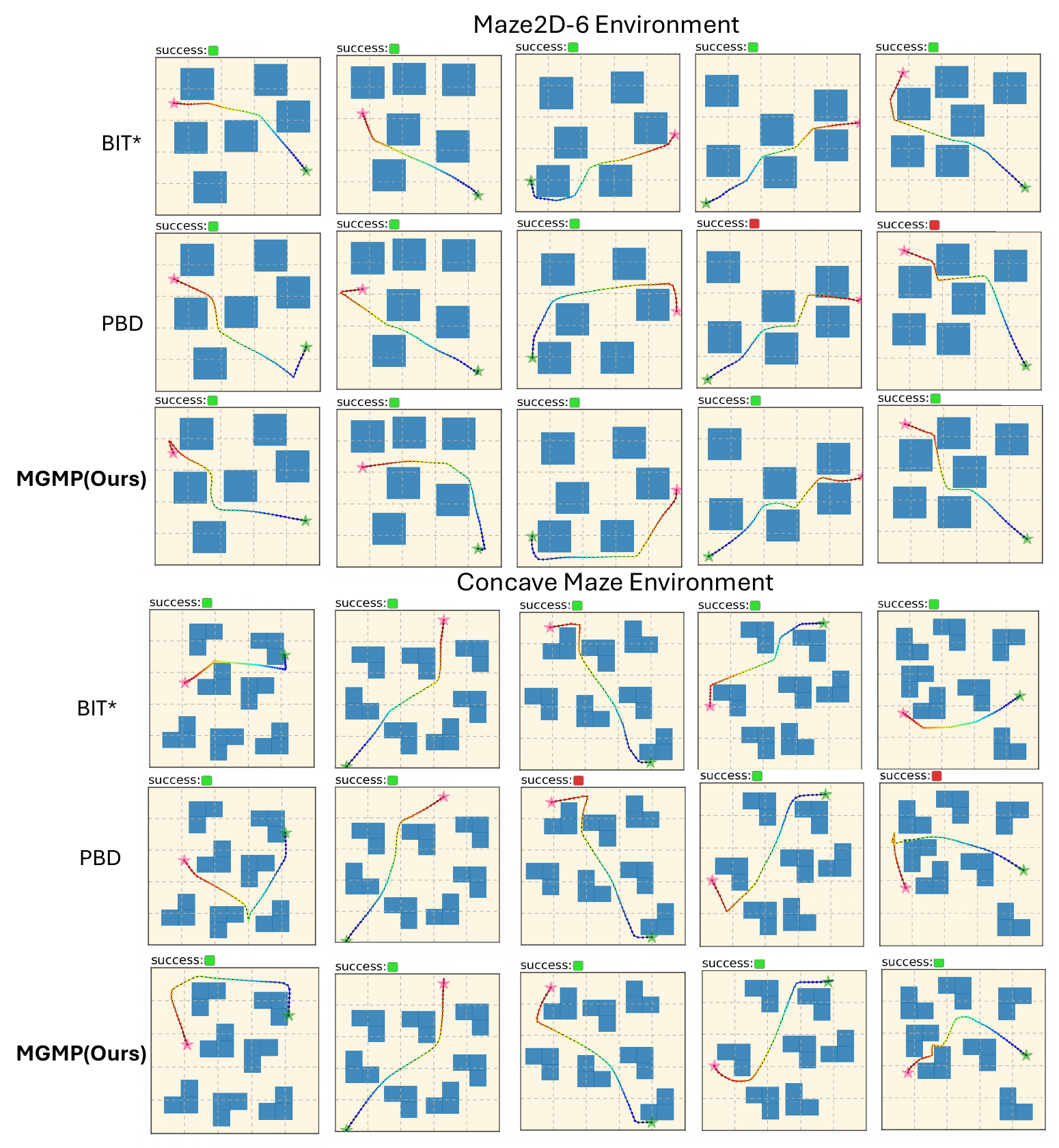}
   \caption{Qualitative results on unseen obstacle layouts in Maze2D. We compare BIT*, PBD, and MGMP on Maze2D-6 and Concave Maze2D. MGMP generates collision-free trajectories that reach the goal while avoiding unseen obstacle placements.}
   \label{fig:qual_maze}
\end{figure*}

\begin{figure*}[t]
   \centering
    \includegraphics[width=0.8\linewidth]{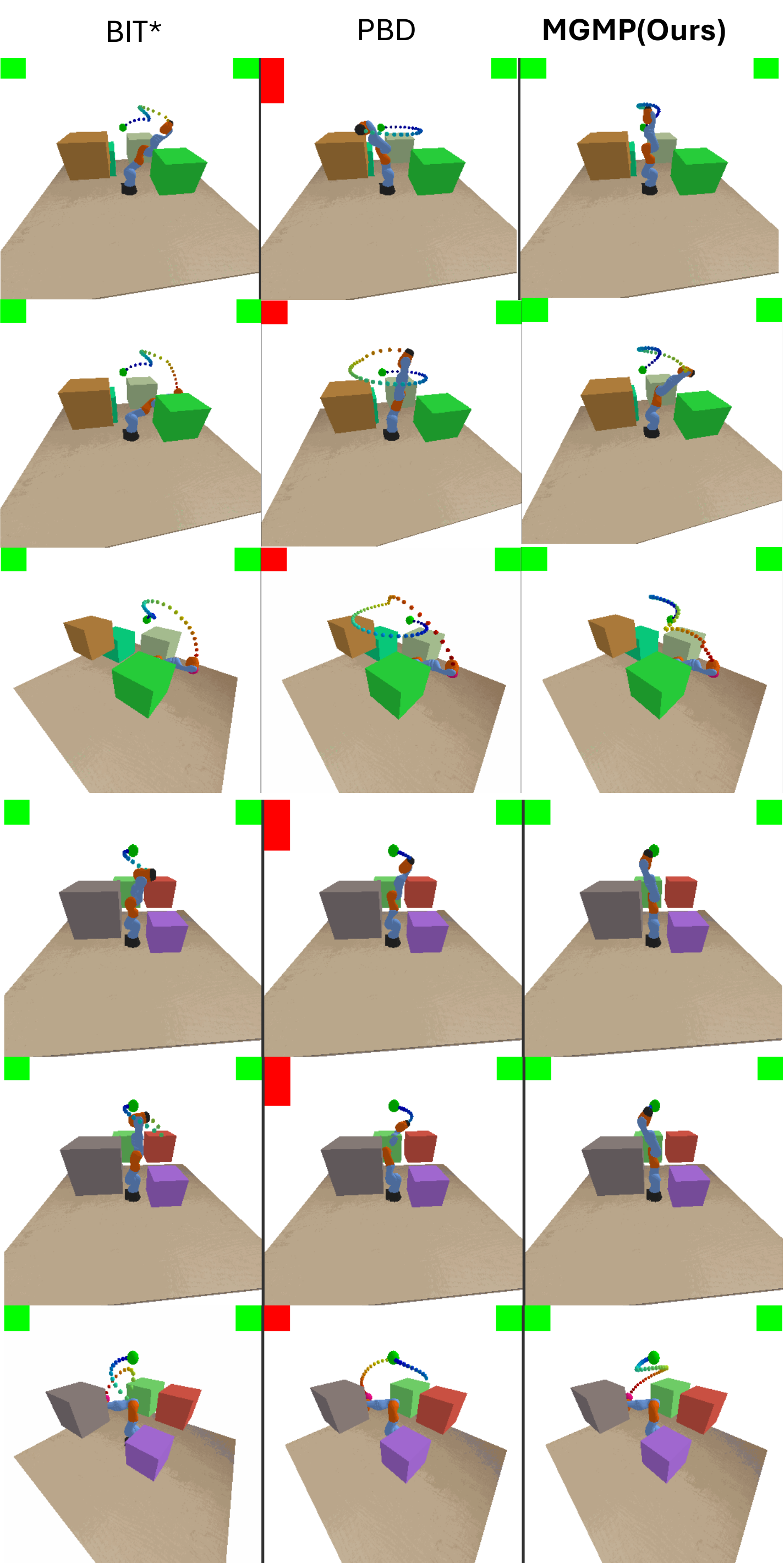}
   \caption{Qualitative results on unseen obstacle layouts in 7-DoF Kuka. Each column compares trajectories generated by BIT*, PBD, and MGMP under the same obstacle configuration. MGMP generates feasible joint-space trajectories that avoid cluttered obstacles.}
   \label{fig:qual_kuka7d_1}
\end{figure*}

\begin{figure*}[t]
   \centering
    \includegraphics[width=0.7\linewidth]{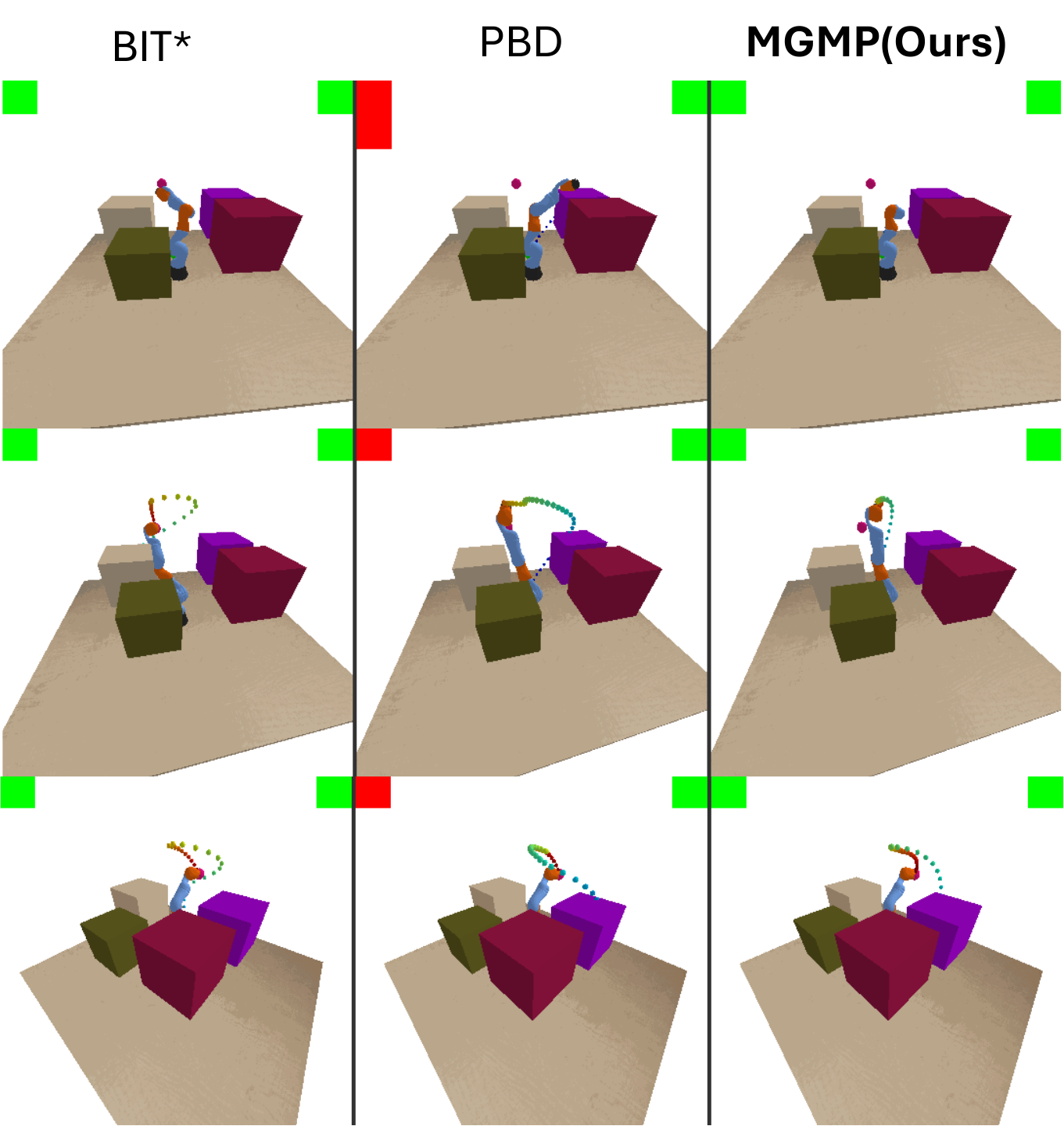}
   \caption{Additional qualitative results on unseen obstacle layouts in 7-DoF Kuka. MGMP produces collision-free plans across different start-goal pairs and obstacle placements.}
   \label{fig:qual_kuka7d_2}
\end{figure*}

\begin{figure*}[t]
   \centering
    \includegraphics[width=0.85\linewidth]{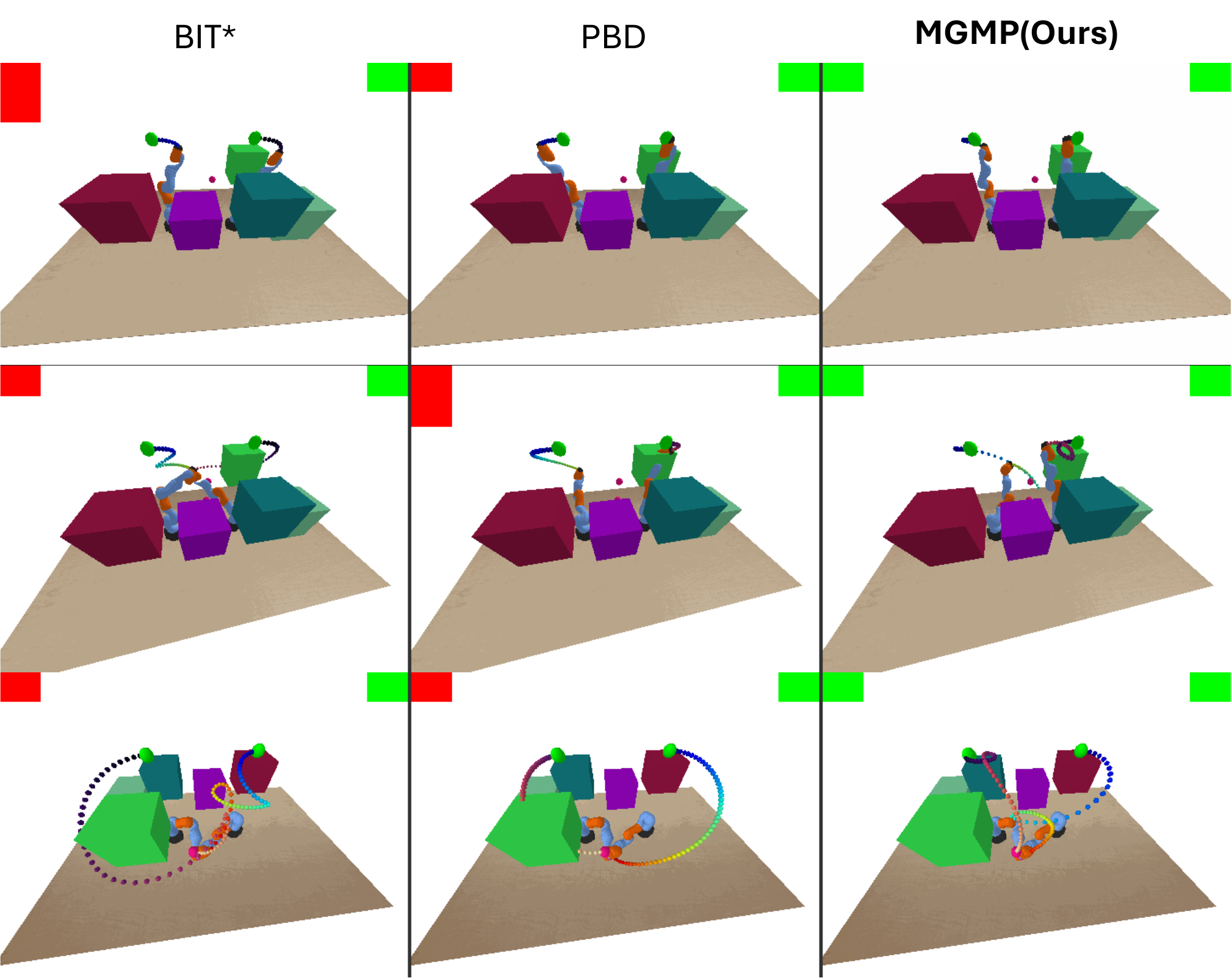}
   \caption{Qualitative results on unseen obstacle layouts in 14-DoF DualKuka. The examples compare BIT*, PBD, and MGMP in higher-dimensional dual-arm planning scenes. MGMP generates feasible trajectories while avoiding both workspace obstacles and inter-arm collisions.}
   \label{fig:qual_dualkuka14d_2}
\end{figure*}

\begin{figure*}[t]
   \centering
    \includegraphics[width=0.85\linewidth]{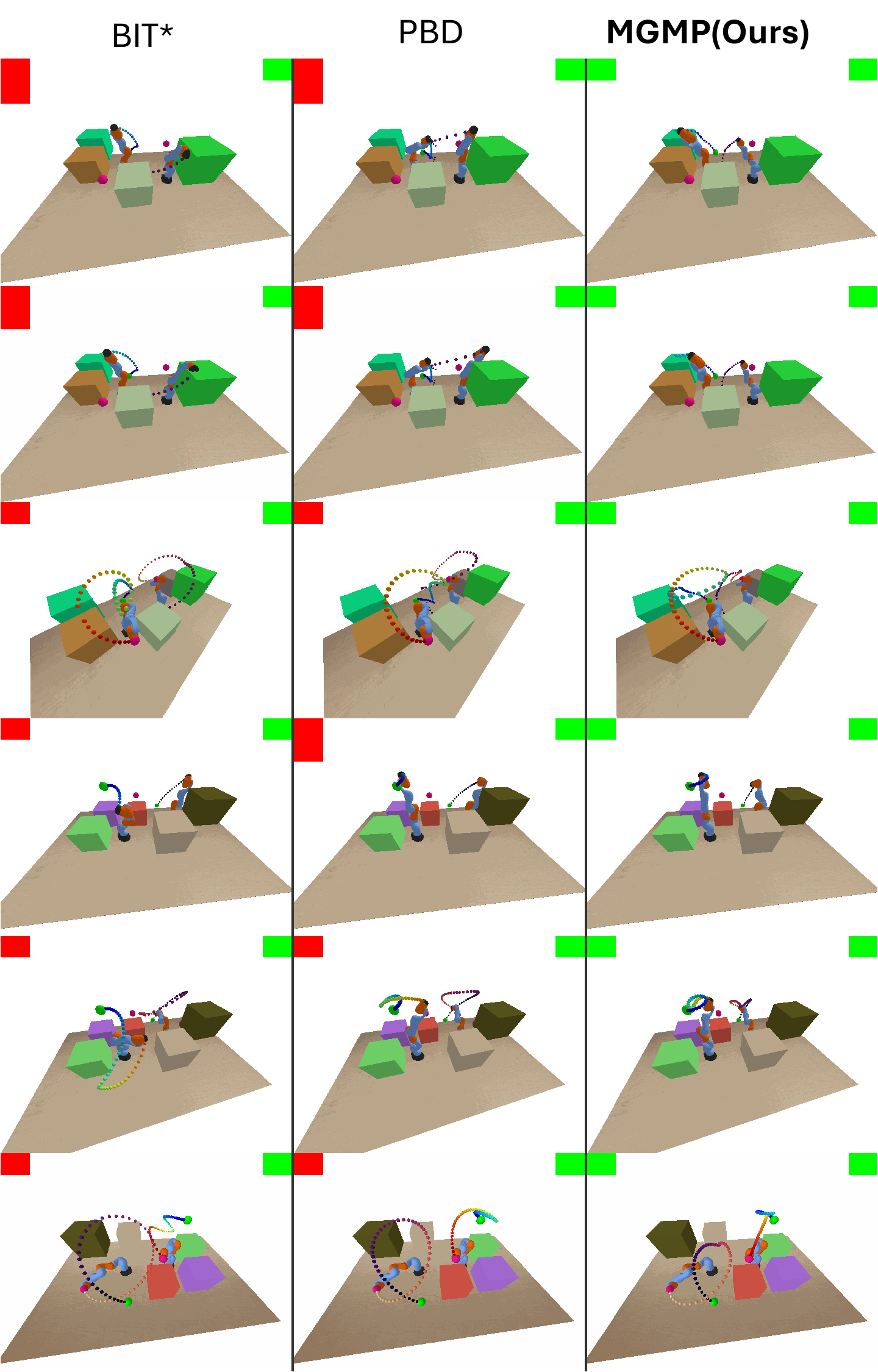}
   \caption{Additional qualitative results on unseen obstacle layouts in 14-DoF DualKuka. MGMP remains robust across different obstacle placements and start-goal configurations.}
   \label{fig:qual_dualkuka14d_1}
\end{figure*}

Figure~\ref{fig:qual_maze} shows qualitative results on the Maze2D environments. Across both Maze2D-6 and Concave Maze2D, MGMP generates collision-free paths that reach the goal while avoiding unseen obstacle layouts. Figures~\ref{fig:qual_kuka7d_1}--\ref{fig:qual_dualkuka14d_1} further show representative Kuka and DualKuka trajectories. MGMP produces feasible joint-space plans in cluttered scenes.

\subsection{Generalization to Additional-Obstacle Environments}\label{sec:Generalization to Additional-Obstacle Environments app}

\begin{figure*}[t]
   \centering
    \includegraphics[width=0.85\linewidth]{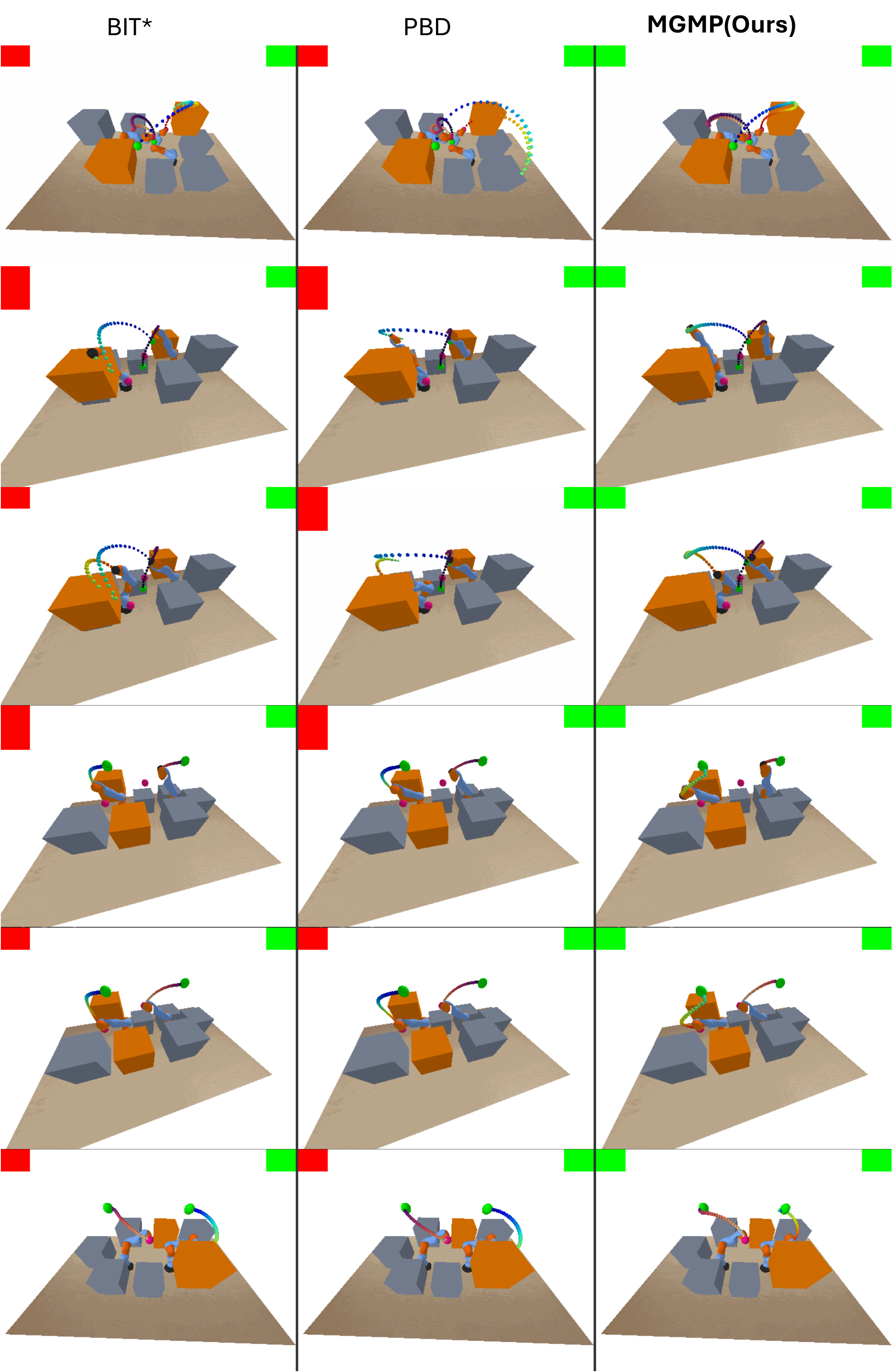}
   \caption{Qualitative results on additional-obstacle 14-DoF DualKuka environments. Compared with the base setting, these scenes include more obstacles at test time. MGMP generates collision-free trajectories in the more constrained workspace.}
   \label{fig:qual_dualkuka14d_add_1}
\end{figure*}

\begin{figure*}[t]
   \centering
    \includegraphics[width=0.85\linewidth]{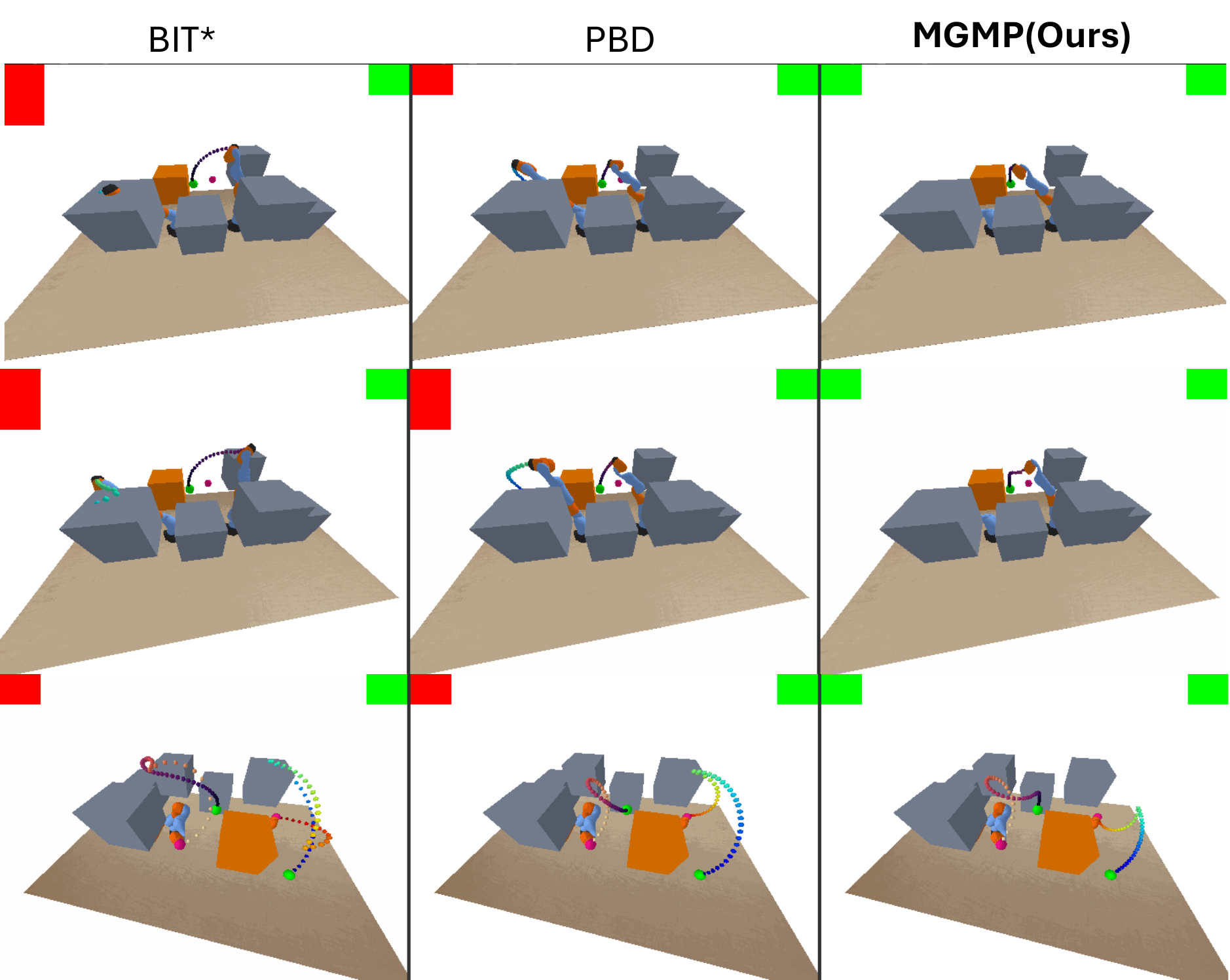}
   \caption{Additional qualitative results on additional-obstacle 14-DoF DualKuka environments. MGMP remains effective when the test scene contains more constraints than those observed during training.}
   \label{fig:qual_dualkuka14d_add_2}
\end{figure*}

Figures~\ref{fig:qual_dualkuka14d_add_1} and~\ref{fig:qual_dualkuka14d_add_2} show qualitative results in additional-obstacle DualKuka environments. These scenes contain more obstacles than those used during training, making collision-free planning more constrained. MGMP still generates feasible trajectories by combining obstacle-conditioned predictions and applying collision-aware refinement.

\subsection{Generalization to Unseen Obstacle Geometries}\label{sec:Generalization to Unseen Obstacle Geometries app}

\begin{figure*}[t]
   \centering
    \includegraphics[width=0.85\linewidth]{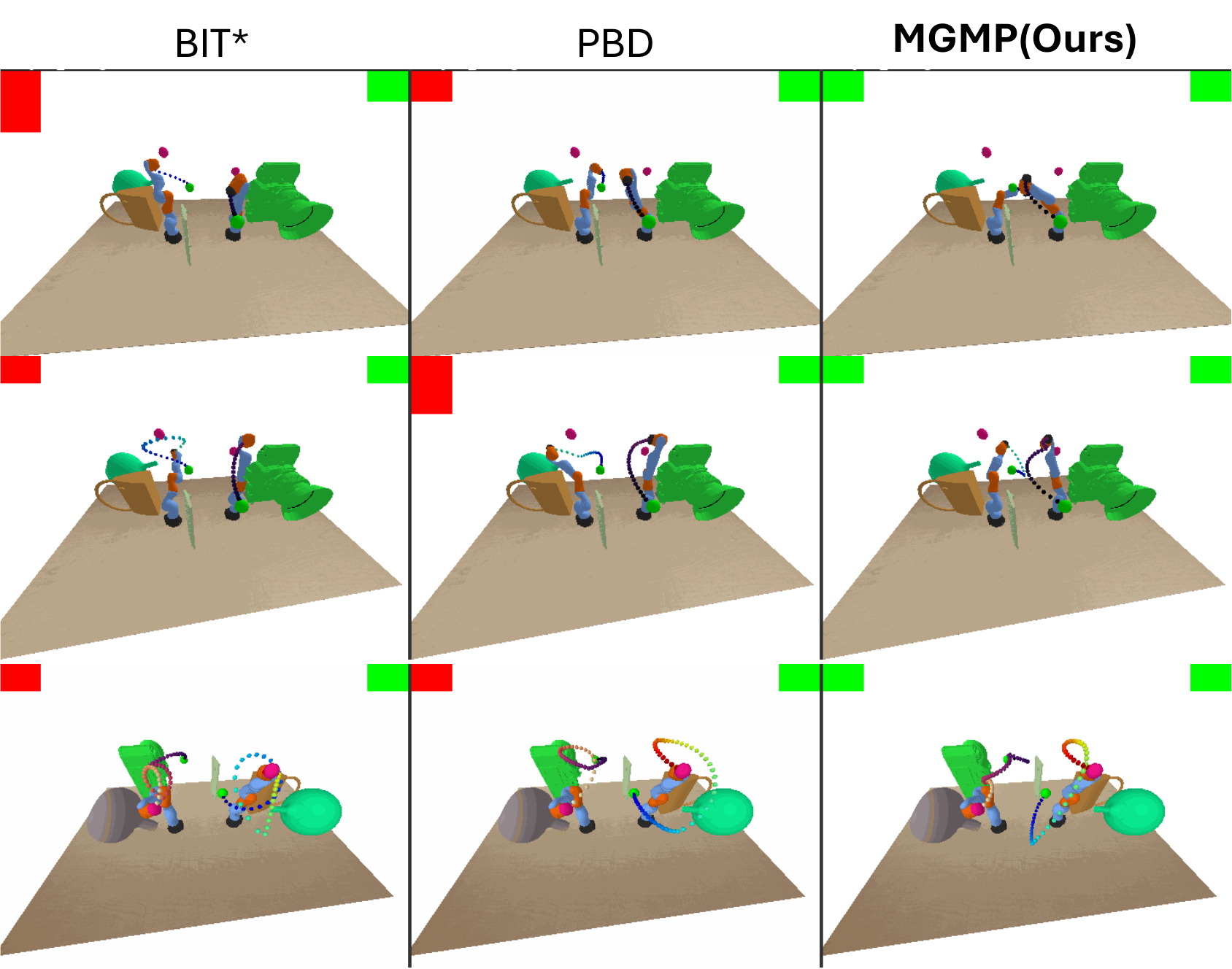}
   \caption{Qualitative results on unseen obstacle geometries in 14-DoF DualKuka. Cuboid training obstacles are replaced by unseen object-level mesh obstacles. MGMP adapts to these geometry shifts and generates collision-free trajectories around irregular objects.}
   \label{fig:qual_dualkuka14d_geo_1}
\end{figure*}

\begin{figure*}[t]
   \centering
    \includegraphics[width=0.85\linewidth]{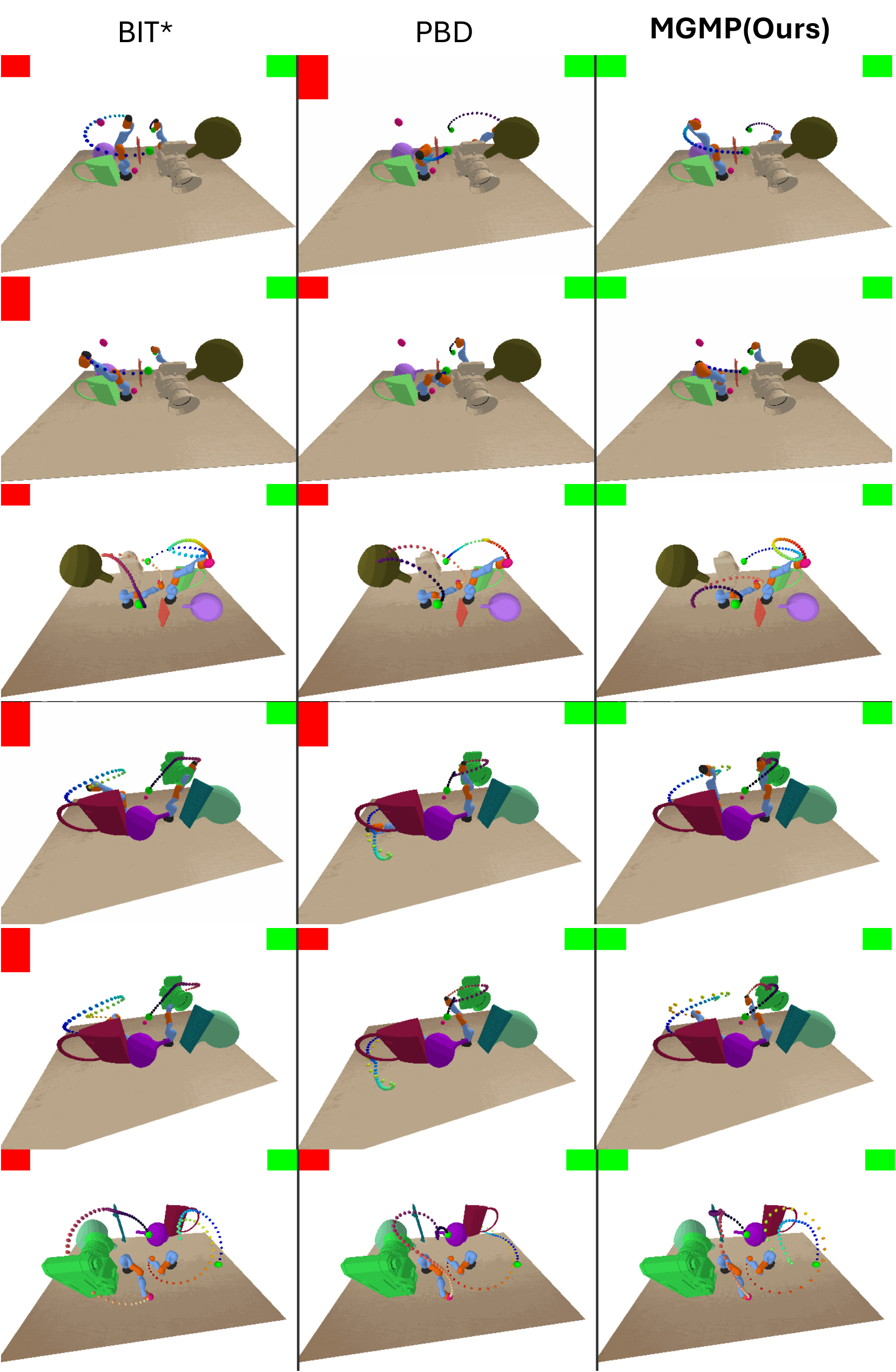}
   \caption{Additional qualitative results on unseen obstacle geometries in 14-DoF DualKuka. MGMP maintains feasible planning performance under realistic mesh obstacles that differ from the box obstacles used during training.}
   \label{fig:qual_dualkuka14d_geo_2}
\end{figure*}

Figures~\ref{fig:qual_dualkuka14d_geo_1} and~\ref{fig:qual_dualkuka14d_geo_2} show qualitative results when box training obstacles are replaced by unseen object-level mesh obstacles. These examples illustrate the ability of MGMP to adapt to irregular obstacle shapes using SDF-guided code refinement. In contrast to the axis-aligned cuboid settings, these environments require the planner to avoid more realistic and non-convex object geometries.

\subsection{Ablation Study And Hyperparameter Experiments}
\label{sec:ablation studyapp}

\begin{table}[t]
\centering
\caption{
Refinement configurations used in the ablation study.
For GGTS variants, discrete refinement consists of position selection,
candidate shortlisting, exact verification, and token commitment,
followed by optional continuous polishing.
}
\label{tab:ggts_design}

\resizebox{0.95\columnwidth}{!}{
\begin{tabular}{lccccc}
\toprule
Variant
&
Position Selection
&
Candidate Shortlist
&
Exact Verification
&
Commit
&
Continuous Opt.
\\
\midrule

MGT (Conf.-Refine)
&
Confidence
&
MGT prob.
&
--
&
--
&
--
\\

MGT + EmbOpt-30
&
--
&
--
&
--
&
--
&
EmbOpt-30
\\

\midrule

Conf.-only Remask
&
Confidence
&
Eq.~(2), $M{\rightarrow}k$
&
top-$k$
&
Sequential
&
Polish
\\

Geom.-only Remask
&
Geometry
&
Eq.~(2), $M{\rightarrow}k$
&
top-$k$
&
Sequential
&
Polish
\\

Probability-only Screening
&
Geom.+Conf.
&
MGT prob., $M{\rightarrow}k$
&
top-$k$
&
Sequential
&
Polish
\\

Independent Exact Search
&
Geom.+Conf.
&
Eq.~(2), $M{\rightarrow}k$
&
top-$k$
&
Independent
&
Polish
\\

MGMP w/o Polish
&
Geom.+Conf.
&
Eq.~(2), $M{\rightarrow}k$
&
top-$k$
&
Sequential
&
--
\\

\textbf{Full MGMP}
&
\textbf{Geom.+Conf.}
&
\textbf{Eq.~(2), $M{\rightarrow}k$}
&
\textbf{top-$k$}
&
\textbf{Sequential}
&
\textbf{Polish}
\\

\bottomrule
\end{tabular}
}

\vspace{0.3em}
\parbox{0.95\columnwidth}{
\footnotesize
For all GGTS variants, token positions associated with detected collision
timesteps are additionally included in the remasking set.
Unless otherwise stated, all variants use the same proposal size $M$,
shortlist size $k$, refinement rounds, and polishing budget.
}
\end{table}

\paragraph{Screening and Exact Verification}
We further analyze the two-stage candidate evaluation used in GGTS.
First-order Only directly commits the top-ranked candidate 
without exact verification, while Exhaustive Verification evaluates all
$M=16$ proposals using the exact objective.
As shown in Table~\ref{tab:ggts_additional}(d), First-order Only reduces
SR/VSF to $0.93/0.78$, indicating that the first-order score is effective
for screening but insufficient as the final discrete decision rule.
Exhaustive Verification achieves $0.98/0.89$ SR/VSF, close to Full GGTS
at $0.97/0.87$, but requires $8\times$ more exact evaluations and
increases runtime from $0.44$\,s to $1.28$\,s.
Thus, the proposed $M\!\rightarrow\!k$ screening retains most of the
benefit of exhaustive verification while substantially reducing
verification cost.

\begin{table*}[t]
\centering
\caption{
Additional analysis of GGTS on 7-DoF Kuka.
(a--c) Sensitivity to the candidate shortlist size, refinement schedule,
and continuous polishing budget.
The MGT proposal size is fixed to $M=16$, and $k$ denotes the number of
candidates retained for exact verification.
For the refinement schedule, $1\times4$, $2\times2$, and $4\times1$
use the same nominal refinement budget, while $1\times1$ serves as a
minimal-refinement reference.
(d) Comparison of first-order screening and exact verification.
Default settings are shown in bold.
}
\label{tab:ggts_additional}

\vspace{0.3em}
\setlength{\tabcolsep}{4.5pt}
\renewcommand{\arraystretch}{1.05}

\resizebox{0.90\textwidth}{!}{
\begin{tabular}{
cccc
@{\hspace{1.5em}}
cccc
@{\hspace{1.5em}}
cccc
}
\toprule

\multicolumn{4}{c}{\textbf{(a) Candidate shortlist size $k$}}
&
\multicolumn{4}{c}{\textbf{(b) Refinement schedule}}
&
\multicolumn{4}{c}{\textbf{(c) Continuous polishing}}
\\

\cmidrule(lr){1-4}
\cmidrule(lr){5-8}
\cmidrule(lr){9-12}

$k$
& SR $\uparrow$
& VSF $\uparrow$
& RT $\downarrow$
&
Schedule
& SR $\uparrow$
& VSF $\uparrow$
& RT $\downarrow$
&
Steps
& SR $\uparrow$
& VSF $\uparrow$
& RT $\downarrow$
\\

\midrule

1
& 0.95 & 0.84 & 0.38
&
$1\times1$
& 0.93 & 0.79 & 0.34
&
0
& 0.94 & 0.83 & 0.40
\\

\textbf{2}
& 0.97 & 0.87 & 0.44
&
$1\times4$
& 0.95 & 0.83 & 0.40
&
3
& 0.96 & 0.86 & 0.42
\\

4
& 0.97 & 0.88 & 0.56
&
$\mathbf{2\times2}$
& 0.97 & 0.87 & 0.44
&
\textbf{5}
& 0.97 & 0.87 & 0.44
\\

16
& 0.98 & 0.89 & 1.28
&
$4\times1$
& 0.96 & 0.85 & 0.52
&
10
& 0.97 & 0.88 & 0.48
\\

\bottomrule
\end{tabular}
}

\vspace{0.8em}

\textbf{(d) Screening and exact verification}\\[0.3em]

\resizebox{0.62\textwidth}{!}{
\begin{tabular}{lcccc}
\toprule
Variant
&
SR $\uparrow$
&
VSF $\uparrow$
&
Exact Evals. $\downarrow$
&
RT $\downarrow$
\\
\midrule

First-order Only
& 0.93 & 0.78 & 0 & 0.32
\\

\textbf{Full GGTS}
& 0.97 & 0.87 & 8 & 0.44
\\

Exhaustive Verification
& 0.98 & 0.89 & 64 & 1.28
\\

\bottomrule
\end{tabular}
}

\end{table*}

\paragraph{Refinement Hyperparameters}

Table~\ref{tab:ggts_additional}(a--c) studies the sensitivity of GGTS to the candidate shortlist size, refinement schedule, and continuous polishing budget on 7-DoF Kuka. For candidate screening, performance improves from $0.95$ SR / $0.84$ VSF at $k=1$ to $0.97/0.87$ at $k=2$, while further increasing $k$ to 4 or 16 provides only marginal gains at substantially higher runtime. We therefore use $k=2$, reducing the $M=16$ MGT proposals to only two
candidates for exact verification.

For the refinement schedule, $2\times2$ achieves the best overall performance among the equal-budget configurations, reaching $0.97$ SR and $0.87$ VSF, compared with $0.95/0.83$ for $1\times4$ and $0.96/0.85$ for $4\times1$.
Finally, continuous polishing improves feasibility from $0.94/0.83$ SR/VSF without polishing to $0.97/0.87$ after five steps, while increasing the budget to ten steps yields only a marginal VSF improvement to $0.88$ with no further gain in SR. We therefore use $k=2$, a $2\times2$ refinement schedule, and five polishing steps in the main experiments.

\section{Real-World Experiments}\label{sec:real-world}
\subsection{Environment setup}

\begin{figure*}[t]
   \centering
    \includegraphics[width=1\linewidth]{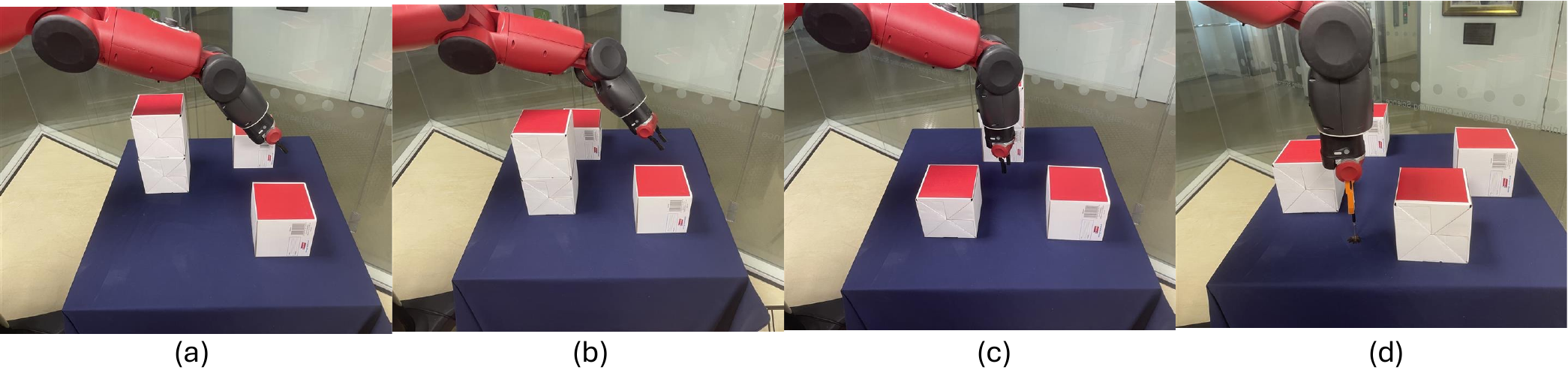}
   \caption{Real-world tabletop setups for Baxter experiments. Figures (a)--(c) show the 3D trajectory setting used for shelf placing and button pressing, where the robot plans obstacle-avoiding motions through the workspace. Panel (d) shows the 2D tabletop drawing setting, where the robot follows a near-surface trajectory while holding a pen.}
   \label{fig:real_setup}
\end{figure*}

We conduct real-world tabletop obstacle-avoidance experiments using the Baxter robot's left arm.  Figure~\ref{fig:real_setup} shows the real-world setups used for data collection. Figures~\ref{fig:real_setup}(a)--(c) show the 3D trajectory setting, where box obstacles are placed on the table and the robot moves through the workspace while avoiding them. This setting is used for shelf placing and button pressing. Figure~\ref{fig:real_setup}(d) shows the 2D tabletop drawing setting, where the robot holds a pen and follows a near-surface trajectory on the table.

\paragraph{Training.} We collect two real-world datasets of task-agnostic trajectories. The 3D dataset contains 200 trajectories across 20 obstacle configurations, with 10 trajectories per configuration. Demonstrations are collected by guiding the arm in zero-gravity mode from a start region to a goal region while avoiding four tabletop box obstacles. Obstacle centers are measured using HTC Vive trackers, and robot joint angles are recorded along each trajectory. The 2D tabletop dataset contains 200 near-surface gripper trajectories across 10 obstacle configurations for pen drawing.

\paragraph{Evaluation on downstream tasks.}
For shelf placing and button pressing, MGMP generates a trajectory to a task-specific goal; the robot follows this trajectory then executes a simple hand-coded primitive action, i.e. releasing the object on the shelf or moving the end-effector downward to press the button. For drawing, the robot grasps a pen and follows the 2D trajectory produced by MGMP to draw a line.

\subsection{Qualitative results}

\begin{figure*}[t]
   \centering
    \includegraphics[width=1\linewidth]{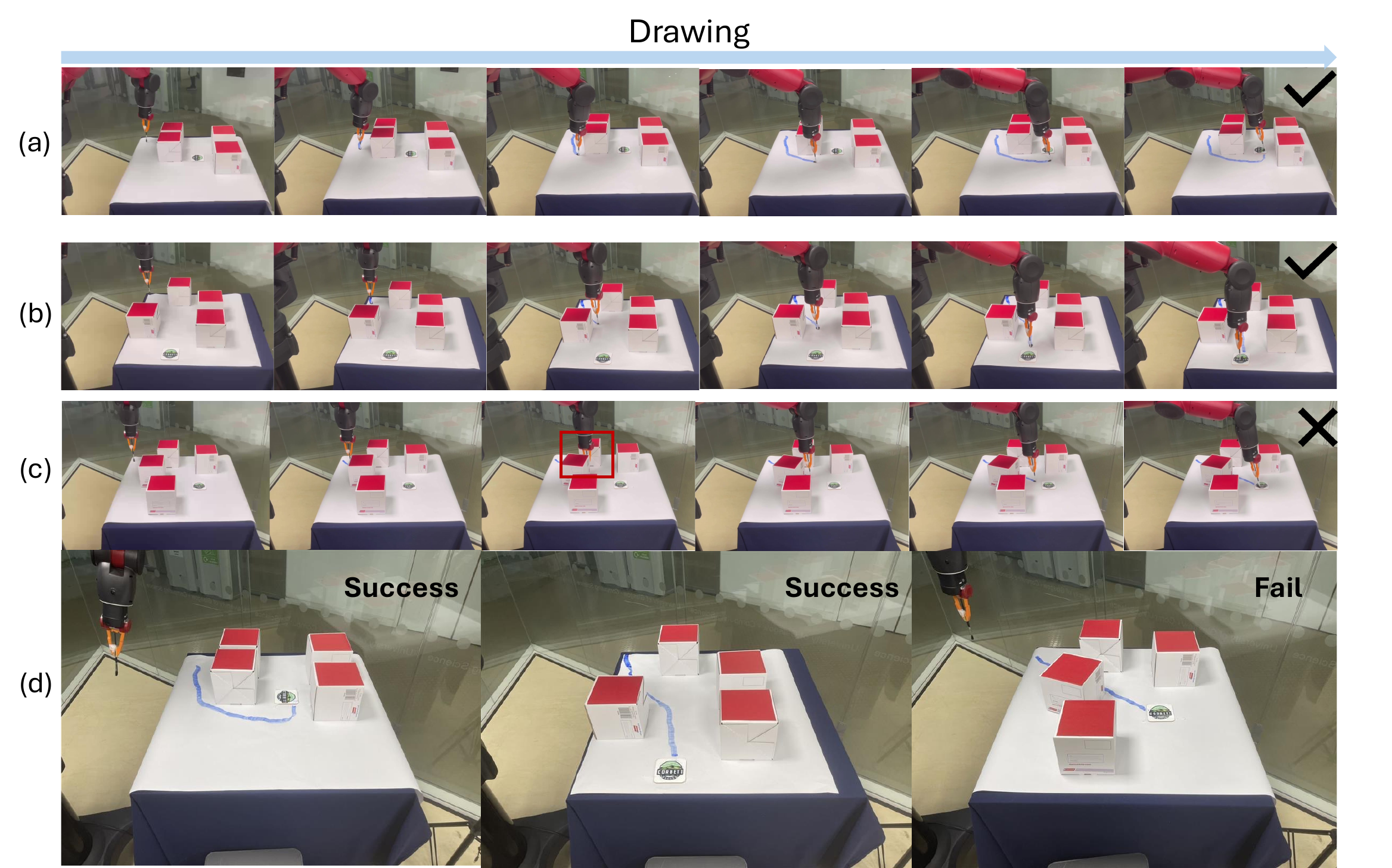}
   \caption{Qualitative results for the real-world pen-drawing task. Rows (a) and (b) show successful executions, where the robot follows the planned 2D tabletop trajectory and draws a line to the target while avoiding obstacles. Row (c) shows a failure case caused by collision with an obstacle during execution. Row (d) shows final drawn trajectories.}
   \label{fig:drawing}
\end{figure*}

\begin{figure*}[t]
   \centering
    \includegraphics[width=0.9\linewidth]{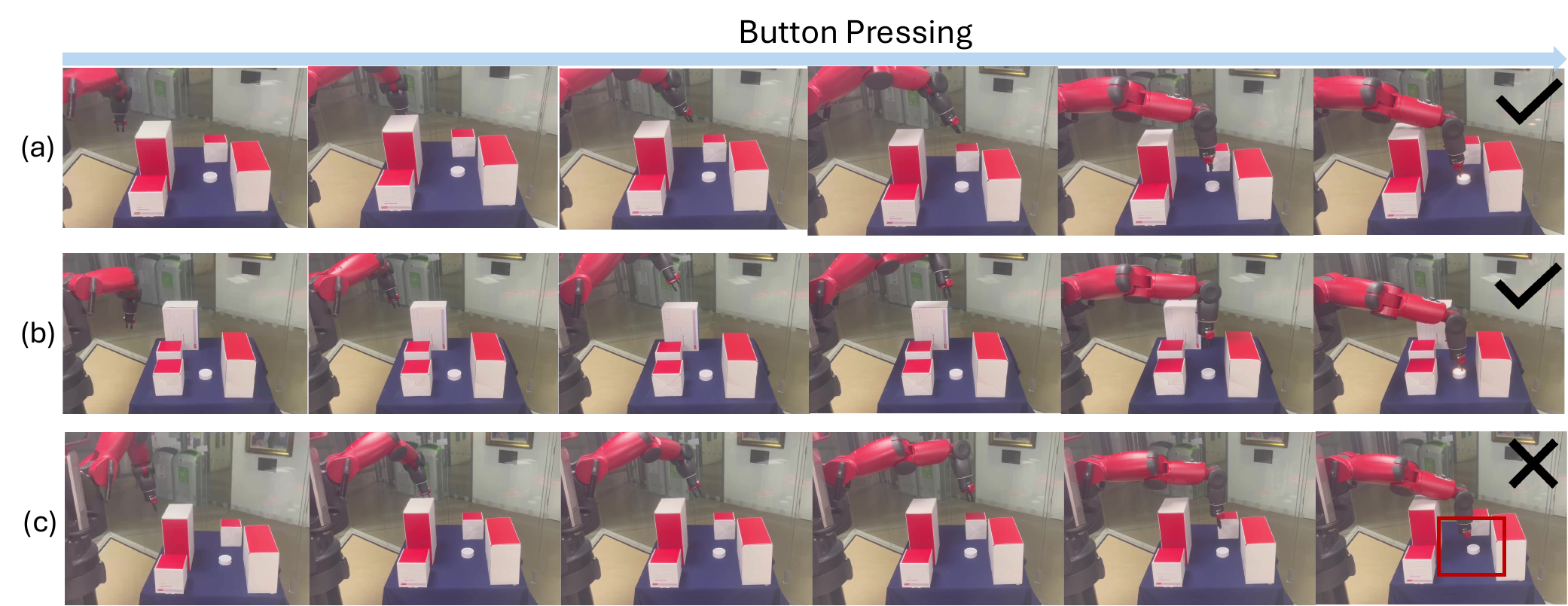}
   \caption{Qualitative results for the real-world button-pressing task. Rows (a) and (b) show successful executions. Row (c) shows a failure case: the robot does not reach the correct pre-press goal pose.}
   \label{fig:button-press}
\end{figure*}

\begin{figure*}[t]
   \centering
    \includegraphics[width=0.95\linewidth]{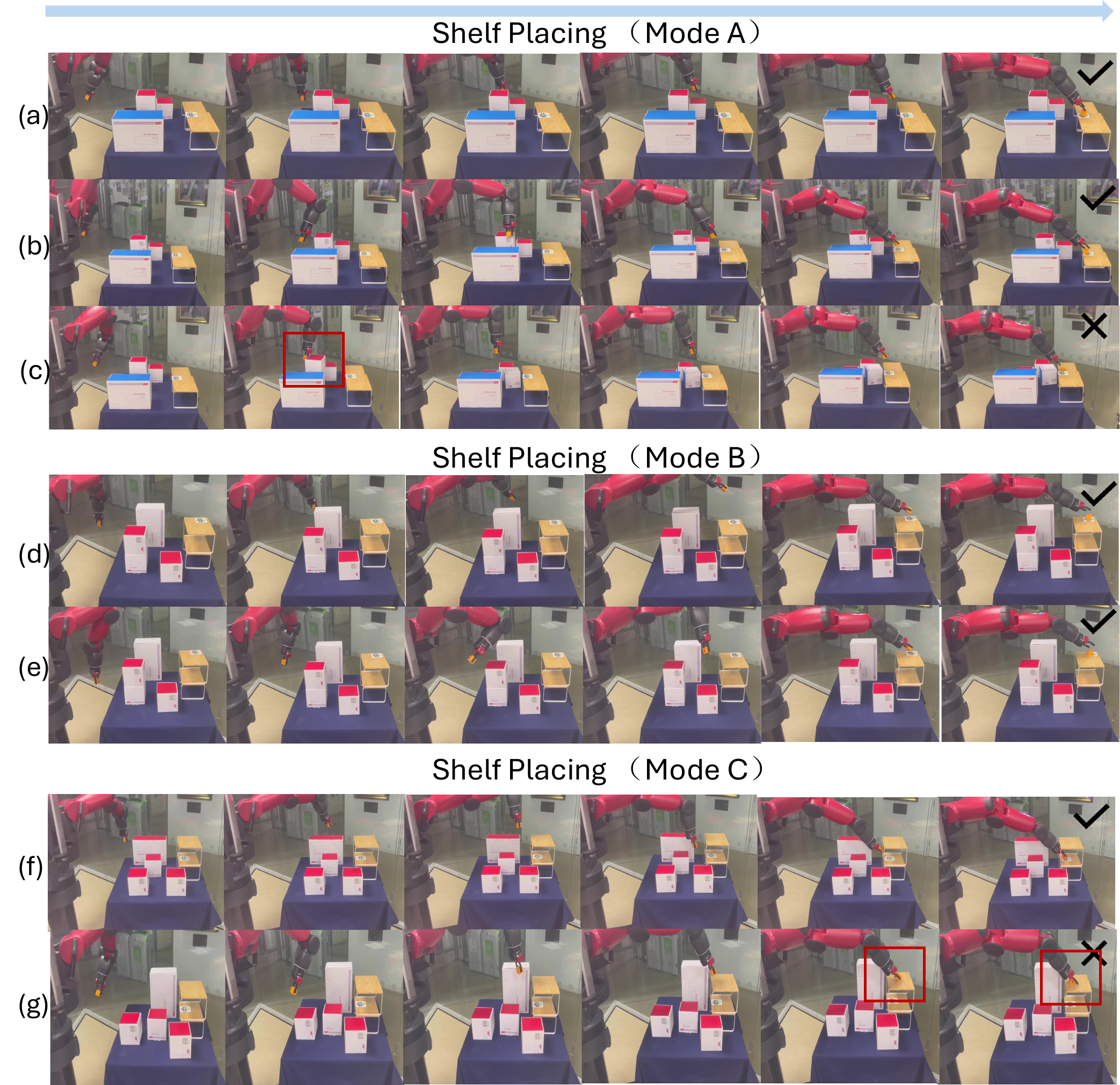}
   \caption{Qualitative results for the real-world shelf-placing task under three target shelf modes. MGMP successfully generates obstacle-avoiding trajectories that reach valid release poses. Failure cases are mainly caused by obstacle collisions or unsuitable approach trajectories that prevent successful placement.}
   \label{fig:shelf-placing}
\end{figure*}

Figures~\ref{fig:drawing}--\ref{fig:shelf-placing} show qualitative results from the real-world experiments. Figure~\ref{fig:drawing} presents the pen-drawing task. In successful cases, the robot follows the planned tabletop trajectory and draws a continuous line toward the target region while avoiding surrounding obstacles. Failures occur when the trajectory is blocked by nearby boxes, leading to obstacle contact or preventing the pen from reaching the desired final location.

Figure~\ref{fig:button-press} shows qualitative results for button pressing. In successful trials, MGMP plans a collision-free approach trajectory that brings the end-effector above the button, after which the downward primitive completes the press. The failure case occurs when the generated trajectory does not place the end-effector at a suitable pre-press pose; as a result, the robot fails to reach the task-specific goal required for activating the button.

Figure~\ref{fig:shelf-placing} shows qualitative results for shelf placing under three modes with different target shelf regions. In successful trials, MGMP generates trajectories that move the object around obstacles and reach valid release poses near the target shelf region. Failures occur when the trajectory collides with a nearby obstacle, as shown in row (c), or when the planned trajectory approaches the shelf from an unsuitable direction and is blocked by the upper shelf layer, as shown in row (g). Rows (b) and (c) use the same environment but different planned trajectories for placing the cube, illustrating the multimodality of MGMP's generated actions.

\end{document}